\documentclass[10pt,twocolumn,letterpaper]{article}

\usepackage[pagenumbers]{cvpr} 

\usepackage{algorithm}
\usepackage{algorithmic}
\usepackage{arydshln}

\newcommand{\grantsponsor}[1]{#1}
\newcommand{\grantnumber}[2]{$^{#1}$#2}  

\definecolor{cvprblue}{rgb}{0.21,0.49,0.74}
\usepackage[pagebackref,breaklinks,colorlinks,allcolors=cvprblue]{hyperref}

\def\paperID{43653} 
\def\confName{CVPR}
\def\confYear{2026}

\title{ADAPT: \underline{A}ttention \underline{D}riven \underline{A}daptive \underline{P}rompt Scheduling and In\underline{T}erpolating Orthogonal Complements for Rare Concepts Generation}

\author{Kwanyoung Lee \quad Hyunwoo Oh \quad SeungJu Cha \quad Sungho Koh \quad Dong-Jin Kim\\
Hanyang University, South Korea\\
{\tt\small \{mobled37, komjii, sju9020, ksh000906, djdkim\}@hanyang.ac.kr}
}

\begin{document}

\twocolumn[{
\renewcommand\twocolumn[1][]{#1}
\maketitle
\begin{center}
     \includegraphics[width=1\textwidth]{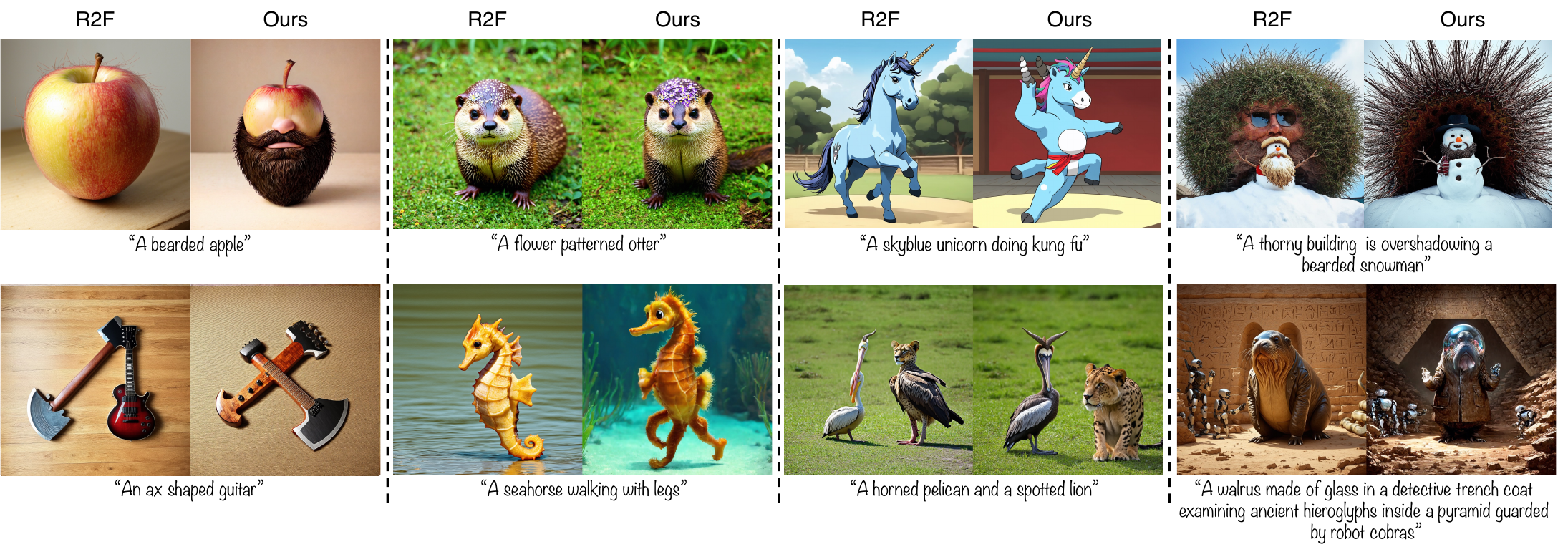}
    \captionof{figure}{Qualitative comparison between R2F and ADAPT (ours) across rare concepts. Our proposed method significantly enhances R2F in a zero-shot way, demonstrating superior capability in text-image alignment. All the pairs are generated with the same seed (42).}
    \label{fig:qualitative}
\end{center}
}]
\begin{abstract}
Generating rare compositional concepts in text-to-image synthesis remains a challenge for diffusion models, particularly for attributes that are uncommon in the training data. 
While recent approaches, such as R2F, address this challenge by utilizing LLM for prompt scheduling, they suffer from inherent variance due to the randomness of language models and suboptimal guidance from iterative text embedding switching.
To address these problems, we propose the ADAPT framework, a training-free framework that deterministically plans and semantically aligns prompt schedules, providing consistent guidance to enhance the composition of rare concepts.
By leveraging attention scores and orthogonal components, ADAPT significantly enhances compositional generation of rare concepts in the RareBench benchmark without additional training or fine-tuning.
Through comprehensive experiments, we demonstrate that ADAPT achieves superior performance in RareBench and accurately reflects the semantic information of rare attributes, providing deterministic and precise control over the generation of rare compositions without compromising visual integrity.
Code is available \href{https://github.com/mobled37/ADAPT}{here}.
\end{abstract}
\section{Introduction}

The generation of rare compositional images has become increasingly important as text-to-image models are widely used to create such compositions~\cite{kirstain2023pick}.
However, generating rare compositional concepts in text-to-image synthesis remains challenging for diffusion models~\cite{esser2024scaling, rombach2022high, podell2023sdxl}, particularly for attributes that are uncommon or absent in training data~\cite{samuel2024generating}. 
Furthermore, existing attribute binding methods~\cite{rassin2023linguistic, chefer2023attend} cannot accurately bind a rare attribute to the common object, such as ``a banana-shaped car'' and ``a black and white checkerboard crocodile.''
R2F~\cite{park2024rare} addresses this by leveraging GPT-4o~\cite{hurst2024gpt} for concept mapping to generate auxiliary frequent prompts and to determine visual detail levels, which linearly map to scheduling stop points between rare and frequent concepts during generation.

However, R2F's dependence on GPT-4o induces variance in the created prompts and visual detail levels due to the inherent randomness of the language model.
Additionally, linear mapping from visual detail levels to stop points is heuristic and misaligned with token-level semantic information during generation.
Furthermore, R2F iteratively switches text embeddings between rare and frequent prompts, making it difficult to provide semantically precise and consistent guidance during the generation process.
In this work, our goal is to utilize deterministic prompt schedules with semantically precise and consistent guidance through denoising steps for generating rare compositions.

To address the problem of variance, we propose \emph{Adaptive Prompt Scheduling (APS)} to mitigate GPT-4o variance on visual detail levels and semantically align prompt scheduling with text tokens.
In particular, we leverage max spatial attention scores to measure attention activation for each token during generation~\cite{chefer2023attend}.
Our key insight leverages spatial attention convergence~\cite{yi2024towards} as a principled indicator of semantic feature saturation, thereby enabling semantically aligned determination of optimal stop points for prompt scheduling.

Second, to improve guidance for rare and frequent concepts, we propose \emph{Pooled Embedding Manipulation (PEM)}, which consistently applies a merged pooled embedding between rare and frequent prompts, rather than iteratively switching between them. 
When merging rare and frequent prompts, we propose to project the rare pooled text embedding onto the orthogonal complement of the frequent prompt, thereby extracting a direction that disentangles rare-specific semantics.
We then interpolate between frequent and rare-specific semantics~\cite{dalva2024fluxspace}, but uniform interpolation strength results in over-suppressing base semantics or under-emphasizing rare attributes.
To address this, we introduce an adaptive weighting strategy~\cite{wang2025precise} that determines the interpolation scale based on the cosine similarity of CLIP's pooled embedding space.
This adaptive scale is then used for linear interpolation along the projection directions, yielding modulated pooled embeddings that balance base semantic preservation with the enhancement of rare attributes.
Lastly, some prompts exhibit substantial semantic differences between rare and frequent concepts (e.g., ``A metallic humanoid figure'' and ``A clown made of steel''), making attribute-specific manipulation challenging.
Therefore, we extract the attribute text (e.g., ``made of steel'' in ``A clown made of steel'') by modifying R2F's concept mapping instructions for LLM and introduce \emph{Latent Space Manipulation (LSM)} to extract disentangled guidance from the attribute text and apply it to the model within attention layers through an orthogonal guidance vector with a tunable scaling factor.

Through these contributions, we propose a framework that improves guidance with rare and frequent concepts, \emph{ADAPT}, for generating rare compositional concepts:
\noindent Our contributions are summarized as follows:
\begin{itemize}

    \item We propose \emph{APS}, which leverages attention scores to determine optimal stop points for prompt scheduling, thereby eliminating GPT-4o dependency in determining stop points and achieving semantic alignment.
    
    \item We introduce a dual-level embedding manipulation framework that combines \emph{PEM} for precise and consistent guidance on rare semantics and \emph{LSM} for attribute-specific guidance, enabling disentangled directional control over rare concept generation.
    
    \item Through comprehensive experiments, we demonstrate that our framework achieves superior performance in accurately reflecting rare attribute semantic information during the generation process.
    
\end{itemize}

\section{Related Work}

\subsection{Enhancing Text-Alignment in Diffusion Models}
Recently, diffusion models~\cite{rombach2022high,podell2023sdxl,chen2023pixart,esser2024scaling} have demonstrated impressive capabilities in generating photorealistic images conditioned on diverse textual inputs.
However, they often struggle to fully comprehend complex text prompts, leading to suboptimal or misaligned generations~\cite{rassin2023linguistic}.
To improve text-image alignment, various strategies have been proposed, including the use of additional conditions~\cite{li2023gligen, zhang2023adding, xie2023boxdiff}, LLM-based instruction generation~\cite{feng2023layoutgpt, yang2024mastering, Oh2025catch}, and cross-attention information~\cite{rassin2023linguistic, chefer2023attend, cha2025verbdiff}.
For example, \cite{zhang2023adding, li2023gligen, xie2023boxdiff} employ spatial conditions such as bounding boxes to guide the layout according to the text.
LLM instruction-based approaches~\cite{feng2023layoutgpt, yang2024mastering, Oh2025catch} utilize textual refinements or additional context generated by large language models to improve semantic understanding.
Another line of work~\cite{cha2025verbdiff, rassin2023linguistic, chefer2023attend} leverages cross-attention maps to extract token-wise alignments and enhance fine-grained semantic consistency.

While these methods have addressed various text alignment problems, they primarily focus on generalizable and intuitive cases—those that are relatively easy for humans to imagine or interpret.

\subsection{Rare Concept Generation}
To better utilize diffusion models in creative fields, some users have expressed the need to generate rare concept images that may not exist in the real world~\cite{kirstain2023pick}.
Although methods like SynGen~\cite{rassin2023linguistic} and Attend $\&$ Excite~\cite{chefer2023attend} can generate reasonably accurate images for certain rare concepts, they still struggle when the target concept is extremely rare (e.g., ``A horned bearded spotted raccoon smiling'')~\cite{park2024rare}.
To address this issue, researchers have proposed more generalized approaches~\cite{samuel2024generating, park2024rare}.
For example,~\cite{samuel2024generating} shows that rare concepts can be successfully generated by carefully selecting generation seeds in the noise space, using a small set of reference images.
More recently, R2F~\cite{park2024rare} improves the ability of diffusion models to generate rare compositional concepts in a training-free manner by leveraging responses from LLMs.
Specifically, they construct an auxiliary frequent prompt from a rare concept and use it iteratively as input, guided by GPT-4o~\cite{hurst2024gpt} responses.
However, the inherent response variance of GPT-4o causes significant variation in the generated images, limiting the method’s applicability.

Unlike R2F, we focus on producing more stable image outputs by addressing this response variance through attention-based scoring.

\section{Method}
Our method leverages attention maps, pooled text embeddings, and latent space within the multi-modal diffusion transformers (MM-DiT) framework through three key components: (1) Adaptive Prompt Scheduling (APS) approach for deterministic and semantically aligned temporal control, (2) Pooled Embedding Manipulation (PEM) for consistent and improved guidance of both rare and frequent prompts, and (3) Latent Space Manipulation (LSM) for injecting attribute-specific guidance.

\begin{figure*}[t]
    \centering
    \includegraphics[width=0.9\textwidth]{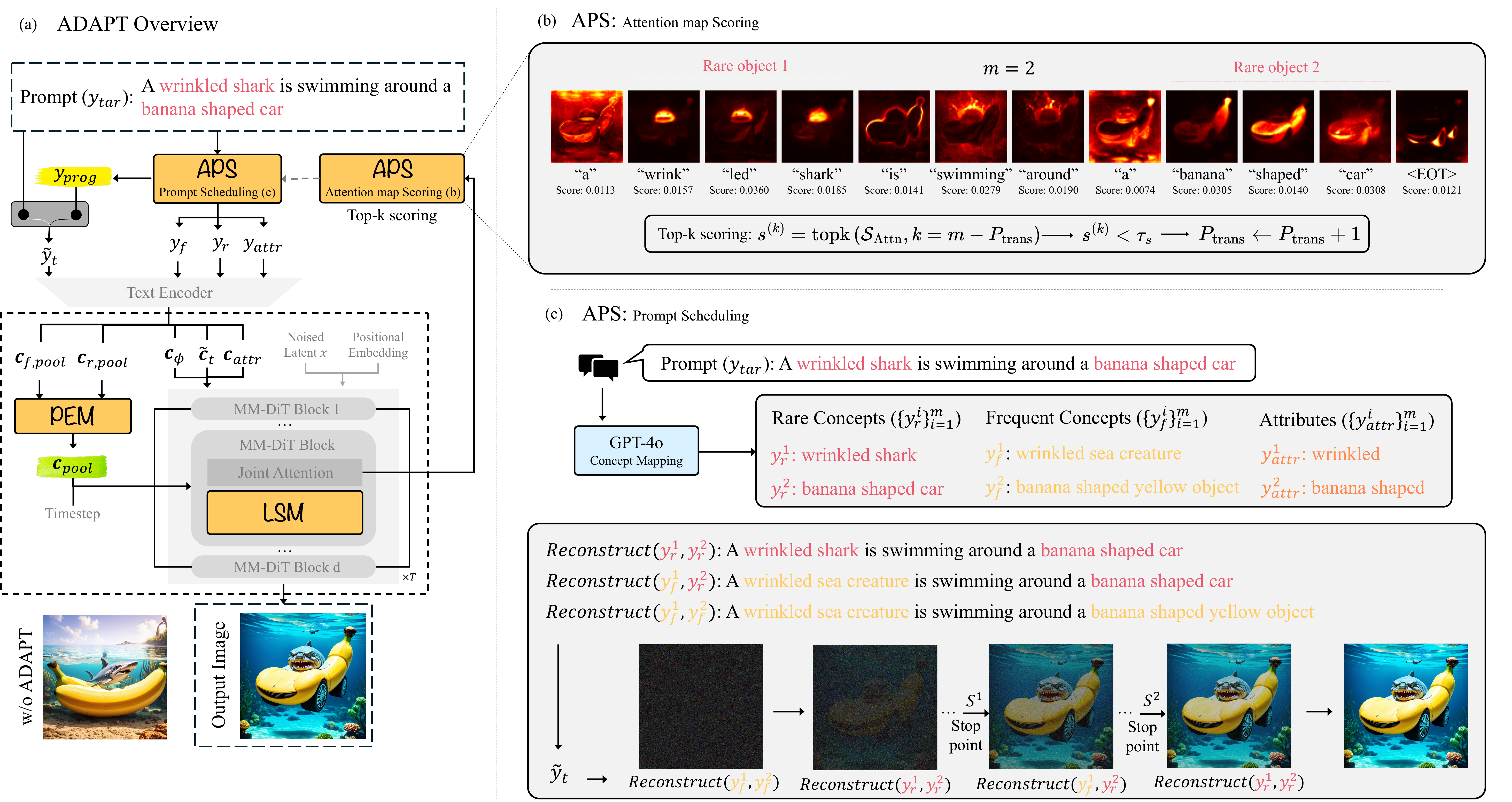}
    \caption{
    Overview of our ADAPT framework.
    (a) Our framework introduces three complementary zero-shot control modules in the Stable Diffusion 3 (SD3) architecture. \emph{Adaptive Prompt Scheduling (APS)} determines optimal stop points $S^i$ based on each token's spatial attention map score $\mathcal{S}_{\text{Attn}}$ at each step; \emph{Pooled Embedding Manipulation (PEM)} operates on the CLIP's pooled text embedding prior to modulation in the SD3 pipeline; \emph{Latent Space Manipulation (LSM)} is applied at the feature level directly within the transformer block after attention computation.
    (b) The overview of attention scoring. 
    We compute Top-k attention response scores $s^{(k)}$ to determine when each token's semantic content has been sufficiently established.
    (c) Visualization of adaptive prompt scheduling between $y_\text{prog}$ and target prompt $y_\text{tar}$ based on stop points $S^i$.
    }
    \label{fig:architecture}
\end{figure*}

\subsection{Preliminaries}
\noindent \textbf{Attention layer.} In multi-modal diffusion transformers, such as MM-DiT~\cite{esser2024scaling}, text-image correspondence is achieved through the attention mechanism within the unified transformer blocks, which enables text-conditioned guidance during generation. 
The pre-trained CLIP and T5 text encoder~\cite{radford2021learning, raffel2020exploring} processes the text prompt $y$ and transforms it into a sequential embedding. 
Similarly, the noised latent $x_t$ is patchified and processed through positional embedding to create image token embeddings.
These text and image embeddings are concatenated into a unified sequence and fed into the MM-DiT blocks. Within each MM-DiT block, the attention mechanism computes relationships across both modalities. The attention map $\mathbf{A}$ is calculated as follows:
\begin{equation}
\mathbf{A} = \text{softmax}\left(\frac{\mathbf{Q}\mathbf{K}^T}{\sqrt{d}}\right),
\label{eq:attention}
\end{equation}
where $\mathbf{Q}$ and $\mathbf{K}$ are derived from the concatenated sequence of text and image tokens, and $d$ represents the channel dimension. For clarity, we omit the denoising timestep $t$ from this notation. 
We denote the spatial attention map for the $i$-th token as $\mathbf{A}^{c}_{y_i}$, where $c$ indicates the spatial map.

\noindent \textbf{Prompt Scheduling.}
R2F~\cite{park2024rare} constructs $m$ number of concept pairs $\{(y_f^i, y_r^i)\}_{i=1}^m$ from a single target prompt $y_{\text{tar}}$, where $y_r^i$ denotes rare prompts (e.g., ``a hairy frog'') and $y_f^i$ are frequent prompts derived from $y_r^i$ using GPT-4o~\cite{hurst2024gpt} (e.g., ``a hairy animal''). 
R2F has proposed a prompt scheduling method by measuring integer visual detail scores $V^i \in \{1,\dots,5\}$ using GPT-4o and normalizing the score values in $\alpha^i \in [0.9, 0.8, 0.6, 0.4, 0.2]$. 
They iterate between $y_r^i$ and $y_f^i$ until the stop point $S^i = \alpha^i T$ (e.g., $S^i = 0.9T$ for $V^i = 1$) diffusion steps, and only use $y_r^i$ after the stop point.
However, the stop point using GPT-4o introduces two limitations: (1) variability in GPT-4o's visual detail levels outputs results in inconsistent scheduling for identical prompts, and (2) the fixed stop points fail to adapt to the semantic progression of generation, reducing controllability and reliability.

\begin{algorithm}[t!]
\caption{Adaptive Prompt Scheduling (APS)}
\label{alg:aps}
\begin{algorithmic}[1]
\REQUIRE Original prompt with rare prompts $\{y_r^i\}_{i=1}^m$ and frequent prompts $\{y_f^i\}_{i=1}^m$, attention threshold $\tau_s$, total timesteps $T$
\ENSURE Generated image with enhanced rare concept representation
\STATE Initialize $P_{\text{trans}} \leftarrow 0$
\STATE Initialize $y_{\text{tar}} \leftarrow \operatorname{Reconstruct}(\{y_r^i\}_{i=1}^m)$
\STATE Initialize $y_{\text{prog}} \leftarrow \operatorname{Reconstruct}(\{y_f^i\}_{i=1}^m)$
\FOR{each timestep $t$ from $T$ to $1$}
    \IF{$(T - t) \% 2 = 0$}
        \STATE Calculate $\mathcal{S}_{\text{Attn}}(y_{\text{tar}}) = \{z_i\}_{i=1}^n$ where
        \STATE \hspace{2.5em} $z_i = \max(\mathbf{A}^c_{y_{\text{tar},i}})$
        \IF{$P_{\text{trans}} < m$}
            \STATE $k \leftarrow m - P_{\text{trans}}$ 
            \STATE $s^{(k)} \leftarrow \operatorname{TopK}(\mathcal{S}_{\text{Attn}}(y_{\text{tar}}), k)$ \COMMENT{$k$-th largest score}
            \IF{$s^{(k)} < \tau_s$}
                \STATE $P_{\text{trans}} \leftarrow P_{\text{trans}} + 1$
                \STATE $y_{\text{prog}} \leftarrow \operatorname{Reconstruct}\left(\{y_f^i\}_{i=1}^{m} \right.$
                \STATE \hspace{2.5em} $ \setminus \{y_f^{P_{\text{trans}}}\}\left.\cup \{y_r^{P_{\text{trans}}}\}\right)$
            \ENDIF
        \ELSE
            \STATE $y_{\text{prog}} \leftarrow y_{\text{tar}}$
        \ENDIF
        \STATE $\tilde{y}_t \leftarrow y_{\text{tar}}$
    \ELSE
        \STATE $\tilde{y}_t \leftarrow y_{\text{prog}}$
    \ENDIF
    \STATE Perform diffusion step with prompt $\tilde{y}_t$
\ENDFOR
\RETURN Generated image
\end{algorithmic}
\end{algorithm}

\subsection{Adaptive Prompt Scheduling (APS)}
To address R2F’s variability and limited semantic alignments, we introduce \emph{Adaptive Prompt Scheduling (APS)}—a deterministic and adaptive strategy that transitions from general to specific concepts by leveraging attention dynamics. 
Unlike R2F, which determines stop points $S^i$ via fixed visual detail heuristics, APS adaptively computes $S^i$ based on the attention scores observed during the generation process.

In SD3's~\cite{esser2024scaling} sequence representation branch, two types of reconstructed prompts are alternated: the progressive prompt $y_{\text{prog}}$ that transitions over timesteps, and a target prompt $y_{\text{tar}}$ that contains all rare concepts.
The target prompt is reconstructed as $y_{\text{tar}}=\operatorname{Reconstruct}(\{y_r^i\}_{i=1}^m)$, composed of only rare concept terms. 
The $\operatorname{Reconstruct}$ operation creates a modified prompt by substituting the detected concept words from the original prompt with their corresponding frequent or rare concept mappings, illustrated in Fig.~\ref{fig:architecture} (c). 

When we construct a $y_{\text{prog}}$ to enable smooth transitions from general to specific concepts, we introduce a counter $P_{\text{trans}}$, initialized to 0 and maximum of $m$, which tracks how many frequent concepts have been successfully replaced by their rare counterparts. 
We determine whether to increment $P_{\text{trans}}$ by analyzing attention probabilities, as they directly reflect the model's focus intensity on specific concepts during generation~\cite{chefer2023attend}.

We calculate the attention response score $\mathcal{S}_\text{Attn}(y_\text{tar})=\{z_i\}_{i=1}^n$ for all tokens in the target prompt $y_\text{tar}$ except $\langle SOS \rangle$ token, where $z_i=\operatorname{max}(\mathbf{A}^c_{y_{\text{tar},i}})$ represents the maximum spatial attention for the $i$-th token and $n$ is the number of tokens. 
These scores guide the transition timing for each rare concept during generation.
While spatial attention generally spreads across all tokens as generation progresses~\cite{yi2024towards}, we observed that tokens differentiating rare and frequent concepts (e.g., ``frog'' in ``A hairy animal'' versus ``A hairy frog'') consistently exhibit the slowest convergence, leading us to switch prompts at these saturation points.
We then inspect the attention scores of the rare concepts that have not yet been transitioned.

\begin{table*}[t]
\centering
\resizebox{0.8\textwidth}{!}{
\centering
\begin{tabular}{l|ccccc|ccc|c}
\hline
\multicolumn{1}{c|}{{}} & \multicolumn{5}{c|}{Single Object}                                                                             & \multicolumn{3}{c|}{Multi Objects} & \multicolumn{1}{c}{}                                                  \\
\multicolumn{1}{c|}{Models}                        & Property      & Shape         & Texture & Action & \begin{tabular}[c]{@{}c@{}}Complex \end{tabular} & Concat & Relation      & \begin{tabular}[c]{@{}c@{}}Complex\end{tabular} & Avg \\ \hline
SD1.5~\cite{rombach2022high}                                         & 55.0            & 38.8          & 33.8    & 23.1   & 36.9                                                        & 23.1   & 24.4          & 36.3     & 33.9                                                 \\
SDXL~\cite{podell2023sdxl}                                          & 60.0            & 56.9          & 71.3    & 47.5   & 58.1                                                        & 39.4   & 35.0            & 47.5    & 52.0                                                  \\
PixArt~\cite{chen2023pixart}                                       & 49.4          & 58.8          & 76.9    & 56.3   & 63.1                                                        & 35.6   & 30.0            & 48.1         &52.3                                             \\
SD3.0~\cite{esser2024scaling}                                        & 49.4          & 76.3          & 53.1    & 71.9   & 65.0                                                          & 55.0     & 51.2          & 70.0            & 61.5                                            \\
FLUX~\cite{blackforestlabs2024flux}                                     & 58.1          & 71.9          & 47.5    & 52.5   & 60.0                                                         & 55.0     & 48.1          & 70.3                 &57.9                                     \\ \hline
SynGen~\cite{rassin2023linguistic}                                       & 61.3          & 59.4          & 54.4    & 33.8   & 50.6                                                        & 30.6   & 33.1          & 29.4                   & 44.1                                   \\ \hline
LMD~\cite{lian2023llm}                                          & 23.8          & 35.6          & 27.5    & 23.8   & 35.6                                                        & 33.1   & 34.4          & 33.1   & 30.9                                                   \\
RPG~\cite{yang2024mastering}                                          & 33.8          & 54.4          & 66.3    & 31.9   & 37.5                                                        & 21.9   & 15.6          & 29.4   & 36.4                                                   \\
ELLA~\cite{hu2024ella}                                         & 31.3          & 61.6          & 64.4    & 43.1   & 66.3                                                        & 42.5   & 50.6          & 51.9                & 51.5                                      \\

R2F (SD3)~\cite{park2024rare}                                    & \underbar{89.4} & \underbar{79.4}    & \underbar{81.9}    & \underbar{80.0}     & \underbar{72.5}         & \underbar{70.0}     & \underbar{58.8}          & \underbar{73.8} & \underbar{75.7}\\
\hline
ADAPT (Ours)  & \textbf{96.3} & \textbf{88.8} & \textbf{83.8} & \textbf{81.9} & \textbf{79.4} & \textbf{76.9} & \textbf{75.0} & \textbf{82.5} & \textbf{83.1}

\\
\hline
\end{tabular}
}
\caption{Text-to-image alignment performances in the RareBench with other baselines with GPT-4o based evaluation. Best values are notated with \textbf{bold}, second-best with \underline{underlined}.}
\label{tab:rarebench_summary}
\end{table*}

\begin{table*}[t!]
\centering
\resizebox{0.8\linewidth}{!}{
\begin{tabular}{l|ccccc|ccc|c}
\hline
\multicolumn{1}{c|}{} & \multicolumn{5}{c|}{Single}                    & \multicolumn{3}{c|}{Multi}  &\multicolumn{1}{c}{} \\
\multicolumn{1}{c|}{Method}                        & Property & Shape & Texture & Action & Complex & Concat & Relation & Complex & Avg\\ \hline
R2F (SD3)                                   & 89.4 & 79.4    & \underbar{81.9}    & \underbar{80.0}     & 72.5         & 70.0     & 58.8          & 73.8            & 75.7                                          \\
\hline

PEM (w/o Adaptive)                    &  90.0        &  84.4     &    78.8     &  71.9      &   76.9      &  71.9      &     73.8     &     79.4  & 78.4  \\
PEM                                       &  \underline{92.5}        &  \textbf{91.3}     &    78.8     &  69.4      &   \underline{78.1}      &  70.6      &     \textbf{77.5}     &     \underline{80.0}   & 79.8 \\
PEM + LSM                                    & \underline{92.5}     & \textbf{91.3}  & 80.6    & 71.9   & 77.5    & 73.1   & \underline{76.9}     & 79.4  & 80.4  \\ 
PEM + APS                                    & \textbf{96.3}     & \underline{88.8}  & 80.0    & 77.5   & \underline{78.1}    & \underbar{74.4}   & 71.3     & 79.4   & \underline{80.7} \\
\hline
PEM + LSM + APS (Ours)                             & \textbf{96.3}     & \underline{88.8}  & \textbf{83.8}    & \textbf{81.9}   & \textbf{79.4}    & \textbf{76.9}   & 75.0 & \textbf{82.5} & \textbf{83.1}   \\ \hline
\end{tabular}
}
\caption{Ablation studies demonstrating the incremental contribution of each proposed component (PEM, LSM, and APS) to text-to-image alignment performance across different evaluation categories on the RareBench dataset.}
\label{tab:ablation_methods}
\end{table*}

Let $k=m-P_{\text{trans}}$ be the number of remaining concepts to transition.
We use the top-$k$ attention scores from $\mathcal{S}_{\text{Attn}}(y_{\text{tar}})$ as our transition indicator, where $s^{(k)}$ denotes the $k$-th largest score in this set.
This ranking-based strategy ensures that we transition the next concept only when at least $k$ concepts have achieved sufficient attention convergence, naturally ordering transitions from most to least saturated concepts.
When $s^{(k)}$ falls below threshold $\tau_s$, indicating that the top-$k$ concepts have sufficiently converged, we trigger a transition: we update $P_\text{trans} \leftarrow P_\text{trans} + 1$ and replace the $P_\text{trans}$-th frequent concept with its rare counterpart:
\begin{equation}
y_{\text{prog}} = \operatorname{Reconstruct}\left(\{y_f^i\}_{i=1}^{m} \setminus \{y_f^{P_{\text{trans}}}\}\cup \{y_r^{P_{\text{trans}}}\}\right).
\end{equation}
As $P_{\text{trans}}$ increases, $k$ decreases, progressively raising the convergence requirement for subsequent transitions.
This adaptive mechanism contrasts with R2F's fixed stop points $S^i = \alpha^i T$, which cannot respond to token-level attention dynamics during generation.

After constructing $y_{\text{prog}}$, at each timestep $t$ of the diffusion process, we alternate these two types of reconstructed prompts:
\begin{equation}
\tilde{y}_t = 
\begin{cases}
y_{\text{prog}}, & \text{if } (T - t) \% 2 \neq 0 \\
y_{\text{tar}}, & \text{if } (T - t) \% 2 = 0
\end{cases}\quad.
\end{equation}
Once all concepts have been transitioned $(P_{\text{trans}} = m)$, the progressive prompt is locked to the target prompt $y_{\text{prog}}=y_{\text{tar}}$.
The overall procedure is summarized in Alg.~\ref{alg:aps}, and a detailed explanation is provided in Appendix~\ref{sec:Algorithm Pseudocode for Adaptive Prompt Scheduling}.

\begin{figure*}[t]
\centering 
\resizebox{0.85\textwidth}{!}{
\centering
     \includegraphics[width=\textwidth]{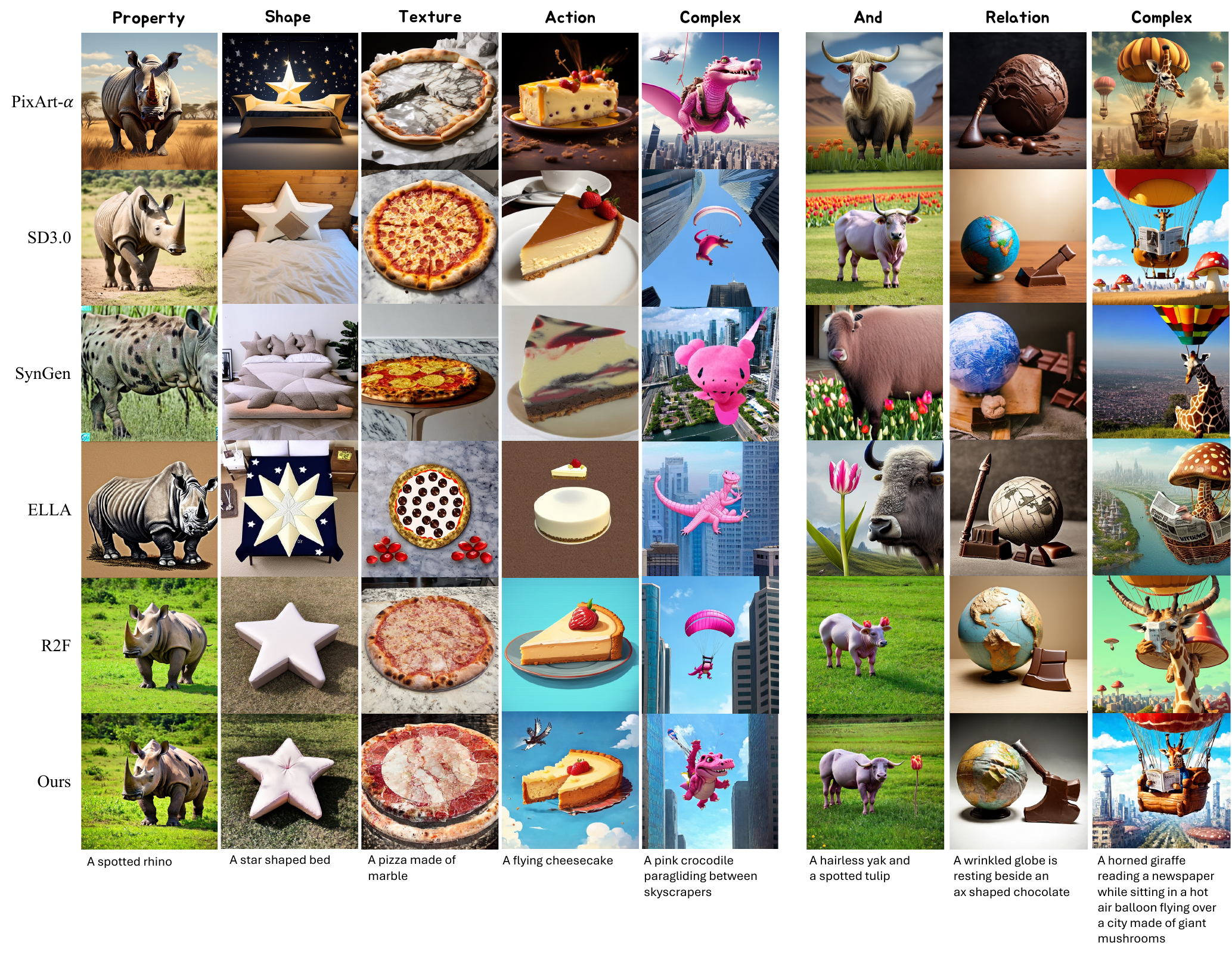}
}
\captionof{figure}{
Qualitative comparison between previous methods and our method across rare concept prompts from RareBench. 
Our approach significantly improves upon R2F in a zero-shot setting, demonstrating superior alignment between image and text for a wide range of challenging attributes and compositions. 
All samples are generated with the same random seed (42) for fair comparison.}
\label{fig:qualitative}
\end{figure*}

\subsection{Pooled Embedding Manipulation (PEM)}
\label{subsec:coarse-editing}

To improve semantic alignment and ensure more consistent guidance during generation, we employ a single, rare-specific pooled embedding that provides disentangled and stable guidance throughout generation.
We use pooled text embeddings from the CLIP model, where $\boldsymbol{c}_{f,\text{pool}}$ and $\boldsymbol{c}_{r,\text{pool}}$ represent the pooled text embeddings for frequent and rare concepts, respectively. 
The corresponding prompts are reconstructed as $y_f=\operatorname{Reconstruct}(\{y_f^i\}_{i=1}^m)$ and $y_r=\operatorname{Reconstruct}(\{y_r^i\}_{i=1}^m)$.
When combining  $\boldsymbol{c}_{f,\text{pool}}$ and $\boldsymbol{c}_{r,\text{pool}}$, in order to disentangle the frequent and rare concept components, we construct the guidance signal $\Delta_{r}$ by orthogonally projecting the rare embedding onto the frequent embedding to isolate novel semantics unique to the rare prompt:

\begin{equation}
    \Delta_{r} = \boldsymbol{c}_{r,\text{pool}} - \frac{\boldsymbol{c}_{f, \text{pool}} \cdot \boldsymbol{c}_{r,\text{pool}}}{\| \boldsymbol{c}_{f, \text{pool} }\|^2} \cdot \boldsymbol{c}_{f, \text{pool}}.
    \label{eq:coarse_proj}
\end{equation}

In~\cite{dalva2024fluxspace}, they applied orthogonal components to edit specific directions using $\Delta_r$ with a simple linear hyperparameter, which can result in over-suppression of base semantics or insufficient emphasis on rare attributes.
Since cosine similarity in CLIP’s pooled embedding reflects semantic alignment, we follow~\cite{wang2025precise} and adaptively scale the edit strength based on the semantic novelty of the rare prompt, measured by its cosine distance from the base embedding.

We define a cosine-based shift factor $\delta(\gamma)$ as:

\begin{equation}
    \delta(\gamma) = \frac{s}{1 + e^{-p (\gamma - \epsilon)}},
    \label{eq:semantic_shift}
\end{equation}
where $\gamma = \cos(\boldsymbol{c}_{r,\text{pool}}, \boldsymbol{c}_{f,\text{pool}})$ measures angular similarity between rare and frequent embeddings. 
Parameters $s > 0$, $p > 0$, and $\epsilon \in (0, 1)$ control the scaling range, sharpness, and similarity threshold, respectively. 
The final pooled embedding is:
\begin{equation}
    \boldsymbol{c}_{\text{pool}} = (1 - \lambda_\text{pool}) \cdot \boldsymbol{c}_{f,\text{pool}} + \lambda_\text{pool} \cdot \delta(\gamma) \cdot \Delta_r,
    \label{eq:coarse_interp}
\end{equation}
where $\lambda_\text{pool} \in [0, 1]$ adjusts the rare direction's contribution and $\delta(\gamma)$ dynamically modulates influence based on semantic dissimilarity. We set $\lambda_\text{pool} = 0.3$ and $(s, p, \epsilon) = (2.0, 100, 0.93)$ following~\cite{wang2025precise} across all experiments.

\subsection{Latent Space Manipulation (LSM)}
For some samples, the frequent prompt has significantly different meanings from the rare prompt (e.g., ``A metallic humanoid figure'' versus ``A clown made of steel''), which leads to unintended semantic directions. 
To provide attribute-specific directions, we adapt the Fine-Grained Editing method~\cite{dalva2024fluxspace} to work with the SD3 model, enabling attribute-specific direction control in the SD3 latent space.

We extract attribute prompts $\{y_\text{attr}^i\}_{i=1}^{m}$ (e.g., ``made of steel'' in ``A clown made of steel'') by modifying LLM instructions for R2F concept mapping (Appendix~\ref{sec:LLM_Instruction_for_ADAPT}).  
We use $\boldsymbol{c}_{\text{attr}}$ as the text embedding corresponding to prompt $y_\text{attr}^{P_{\text{trans}}'}$, where $P_\text{trans}'=\operatorname{min}(P_\text{trans}+1, m)$.
To isolate the unique influence of the attribute control, we compute the orthogonal component: 
\begin{align}
l'_{\theta}(x_t, \boldsymbol{c}_\text{attr},  t) 
&= l_{\theta}(x_t, \boldsymbol{c}_\text{attr}, t) \nonumber \\
&\quad - \frac{
    l_{\theta}(x_t, \boldsymbol{c}_\text{attr}, t) \cdot l_{\theta}(x_t, \boldsymbol{c}_\phi, t)
}{
    \|l_{\theta}(x_t, \boldsymbol{c}_\phi, t)\|^2
} \,\cdot l_{\theta}(x_t, \boldsymbol{c}_\phi, t),
\label{eq:combined_ortho_comp}
\end{align}
where $l_{\theta}(\cdot)$ is attention layer output and $\boldsymbol{c}_{\phi}$ is null text embedding.
The final guided representation is computed via linear interpolation:
\begin{equation}
\hat{l}_{\theta}(x_t, \tilde{\boldsymbol{c}}_t, \boldsymbol{c}_\text{attr}, t) = l_{\theta}(x_t, \tilde{\boldsymbol{c}}_t, t) + \lambda_\text{attr} \cdot \,l'_{\theta}(x_t, \boldsymbol{c}_\text{attr}, t),
\label{eq:editing_scheme}
\end{equation}
\noindent where $\tilde{\boldsymbol{c}}_t$ is the text embedding of $\tilde{y}_t$ and $\lambda_{\text{attr}}\in [0, 1]$ controls the strength of semantic guidance for target attribute alignment.
We set $\lambda_{\text{attr}}=0.15$ across all experiments.

\begin{table*}[t!]
\centering
\resizebox{0.9\linewidth}{!}{
\begin{tabular}{l|ccccc|ccc|c}
\hline
\multicolumn{1}{c|}{} & \multicolumn{5}{c|}{Single}                    & \multicolumn{3}{c|}{Multi}  &\multicolumn{1}{c}{} \\
\multicolumn{1}{c|}{Method}                        & Property & Shape & Texture & Action & Complex & Concat & Relation & Complex & Avg\\ \hline
Simple $y_\text{tar}$                    &  49.4        &  76.3     &    53.1     &  71.9      &   65.0      &  55.0      &     51.2     &     70.0  &  61.5  \\ 
Simple $y_\text{prog}$                    &  57.5        &  72.5     &    55.0     &  46.3      &   56.7      &  42.5      &     48.8     &     53.1  & 54.1  \\

Gram-Schmidt $y_\text{tar}$ and $y_\text{prog}$                                       &  33.8        &  50.0     &    38.1     &  16.9      &   42.5      &  18.8      &     30.6     &     31.2   & 42.7 \\
\hline
R2F (SD3)                                    & \underbar{89.4} & \underbar{79.4}    & \underbar{81.9}    & \underbar{80.0}     & \underbar{72.5}         & \underbar{70.0}     & \underbar{58.8}          & \underbar{73.8}            & \underbar{75.7}                                          \\
ADAPT (Ours)                             & \textbf{96.3}     & \textbf{88.8}  & \textbf{83.8}    & \textbf{81.9}   & \textbf{79.4}    & \textbf{76.9}   & \textbf{75.0} & \textbf{82.5} & \textbf{83.1}   \\ \hline
\end{tabular}
}
\caption{Ablation study on text embedding strategies for the SD3's sequence representation branch. The top section shows single text embedding methods: using only $y_\text{tar}$, using only $y_\text{prog}$, and orthogonal decomposition of $y_\text{tar}$ with $y_\text{prog}$. The bottom section shows iterative switching text embedding methods.}
\label{tab:single_text_embedding}
\end{table*}

\begin{table*}[t!]
\centering
\resizebox{0.9\linewidth}{!}{
\begin{tabular}{l|ccccc|ccc|c}
\hline
\multicolumn{1}{c|}{} & \multicolumn{5}{c|}{Single}                    & \multicolumn{3}{c|}{Multi}  &\multicolumn{1}{c}{} \\
\multicolumn{1}{c|}{Method}                        & Property & Shape & Texture & Action & Complex & Concat & Relation & Complex & Avg\\ \hline
Noun-Only                                   & \underbar{95.6} & \textbf{93.8}    & \underbar{79.4}    & 78.1     & 73.8         & \underbar{76.9}     & 71.3          & \underbar{78.8}            & 81.0                                          \\
Rare Phrases                    &  \textbf{96.3}       &  \underbar{93.1}     &    77.5     &  \underbar{78.8}      &   \underbar{76.9}      &  \textbf{78.8}      &     \underbar{75.0}     &     \underbar{78.8}  & \underbar{81.9}  \\
All tokens (w $\langle SOS\rangle$)                             & 92.5     & 81.9  & 76.3    & 73.8   & \underbar{76.9}    & 72.5   & \textbf{75.6} & 78.1 & 78.5   \\
All tokens (w/o $\langle SOS\rangle$)                             & \textbf{96.3}     & 88.8  & \textbf{83.8}    & \textbf{81.9}   & \textbf{79.4}    & \underbar{76.9}   & \underbar{75.0} & \textbf{82.5} & \textbf{83.1}   \\ \hline
\end{tabular}
}
\caption{Ablation studies examine different attention score extraction strategies.}
\label{tab:token_experiments}
\end{table*}

\section{Experiments}

\noindent \textbf{Implementation Details.} Our work is specifically tailored for the Stable Diffusion 3.0 (SD3.0) architecture.
During inference, we fix the sampling steps to $T = 50$ and use a consistent random seed $(42)$.
The hyperparameters for each components are set as follows: $\tau_s=0.025$, $\lambda_{\text{pool}} = 0.3$, $(s, p, \epsilon)=(2.0, 100, 0.93)$, and $\lambda_{\text{attr}} = 0.15$.

All experiments are implemented using PyTorch 2.7.0 and executed on a single NVIDIA A6000 GPU.

\noindent \textbf{Evaluation.} We evaluate the text-to-image alignment score by GPT-4o, which was originally proposed by~\cite{park2024rare}, and user study (Appendix~\ref{sec:user-study}).

\subsection{Main Results on RareBench}

We evaluate on RareBench~\cite{park2024rare}, a benchmark assessing image generation quality for rare semantic concepts.
Tab.~\ref{tab:rarebench_summary} shows the text-to-image alignment performance on the RareBench benchmark, comparing our method against a range of baselines using GPT-4o evaluation. 
Numerically, our approach achieves performance gains ranging from $+1.9$ to $+16.2$ over R2F in all categories. 
Notably, we observe significant improvements in categories such as \textit{Single-Object Shape} ($+9.4$), \textit{Multi-Object Relation} ($+16.2$), and \textit{Multi-Object Complex} ($+8.7$), indicating stronger alignment with rare compositional semantics.
Fig.~\ref{fig:qualitative} illustrates the qualitative comparison on RareBench across all categories. 
These results suggest that our method demonstrates superior rare semantic composition while maintaining the realism of the image across diverse prompt scenarios.
Additional qualitative results under varying random seeds are provided in Appendix~\ref{sec:Further Visualization}.

\section{Ablation Study}

\begin{table*}[t!]
\centering
\resizebox{0.8\linewidth}{!}{
\begin{tabular}{l|ccccc|ccc|c}
\hline
\multicolumn{1}{c|}{} & \multicolumn{5}{c|}{Single}                    & \multicolumn{3}{c|}{Multi}  &\multicolumn{1}{c}{} \\
\multicolumn{1}{c|}{Method}                        & Property & Shape & Texture & Action & Complex & Concat & Relation & Complex & Avg\\ \hline
Mean  & 71.9 & 87.5    &  63.7   &   76.3   & 75.0         & 67.5     & 65.1         & 79.4            & 73.3                                          \\

Cumulative      &  94.4        &  80.0     &    80.6     &  78.1      &   81.3      &  67.5      &     68.1     &     80.6  & 78.8  \\
Individual                                &  \textbf{96.3}   &  \textbf{88.8} &  \textbf{83.8}  &  \textbf{81.9}      &  \textbf{79.4}  &  \textbf{76.9}      &    \textbf{75.0 }    &   \textbf{82.5}   & \textbf{83.1} \\
\hline
\end{tabular}
}
\caption{Quantitative results for different token aggregation methods in attention scoring.}
\label{tab:token aggregation}
\end{table*}

\noindent \textbf{Quantitative Comparison of Each Proposed Component.}
Tab.~\ref{tab:ablation_methods} shows the quantitative analysis of ADAPT's components and their various combinations.
Incorporating adaptive weighting into PEM results in a notable performance improvement, demonstrating its importance in balancing interpolation strength.
Introducing LSM yields a moderate improvement, enhancing attribute-specific guidance. 
Furthermore, integrating APS notably boosts generation quality in \textit{Property} and \textit{Action} categories, demonstrating its effectiveness in semantically aligning prompt guidance. 
By integrating all components—PEM, LSM, and APS—we obtain the best overall performance, achieving consistent gains across both single- and multi-object settings, validating the effectiveness of our unified framework.

\noindent \textbf{Comparing Single Text Embedding.} Tab.~\ref{tab:single_text_embedding} compares three single guidance approaches with the iterative switching approaches on the SD3's sequence representation branch. 
(1) simple $y_\text{tar}$, (2) simple $y_\text{prog}$, and (3) Gram-Schmidt $y_\text{tar}$ and $y_\text{prog}$ using orthogonal decomposition and linear interpolation between text embeddings.
Overall, using single-prompt embedding leads to significant performance degradation across all evaluation categories.
Notably, (3) performs considerably worse than even (1), indicating that in the sequence representation branch, iterative prompt guidance is necessary for generating rare compositional concepts.

\noindent \textbf{Attention Score Extraction Strategies.} Tab.~\ref{tab:token_experiments} evaluates different attention extraction strategies applied to text tokens.
We extract nouns and rare phrases by using Spacy as a parser.
Noun-only extracts scores exclusively from noun tokens, rare phrases extracts scores from the identified rare phrases, and all tokens extracts scores from all tokens, either including or excluding the $\langle SOS \rangle$ token.
Both noun-only and rare phrase strategies underperform compared to the all-tokens approach, as they struggle with complex compositional prompts such as ``A lion wearing a detective hat and magnifying glass investigating a crime scene on a pirate ship navigated by penguins.''
Excluding the $\langle SOS \rangle$ token yields better results than including it, since the $\langle SOS \rangle$ token fails to converge as generation progresses.
The attention score behavior of the $\langle SOS \rangle$ token is visualized in Appendix~\ref{sec:why_except_the_sos_token}.

\begin{figure}[t!]
    \centering
    \includegraphics[width=0.9\linewidth]{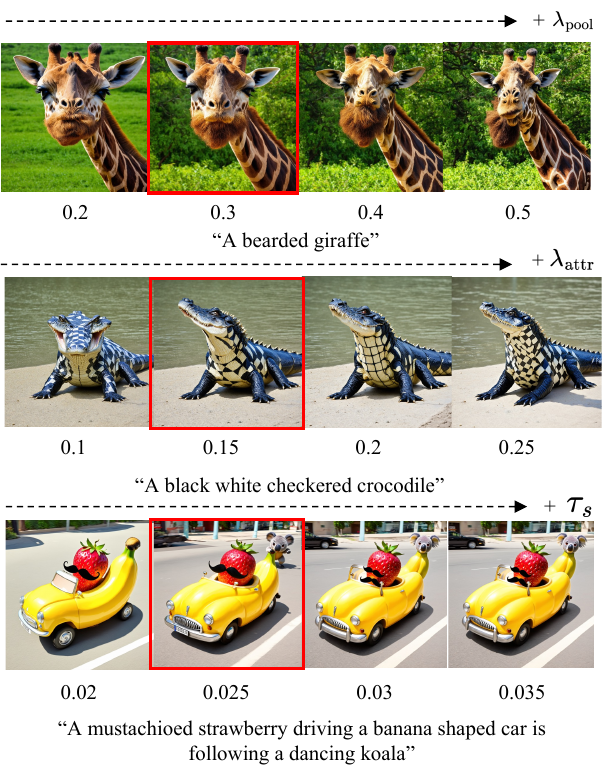}
    \caption{Qualitative comparison on the hyperparameters introduced within the ADAPT framework.}
    \label{fig:ablation_study}
\end{figure}

\noindent \textbf{Impacts on $\lambda_{\text{pool}}$, $\lambda_{\text{attr}}$, and $\tau_s$.} Fig.~\ref{fig:ablation_study} shows hyperparameter impacts on rare attribute generation. 
Increasing $\lambda_{\text{pool}}$ enhances the visibility of rare attributes, but values exceeding $0.5$ degrade visual quality. 
A higher $\lambda_{\text{attr}}$ strengthens the fidelity of attribute and object identity, but values above $0.2$ degrade the fidelity of the crocodile. 
The optimal value $\tau_s=0.025$ enables complete koala generation, whereas $\tau_s=0.02$ fails to produce the koala entirely, and excessive thresholds ($\tau_s=0.03, 0.035$) result in incomplete generation with only koala heads appearing.
Additional visualization results are provided in Appendix~\ref{sec:Further Visualization of Ablation Studies}.

\noindent \textbf{Attention Score Aggregation.}
Tab.~\ref{tab:token aggregation} evaluates different token aggregation strategies for attention scoring. 
Instead of considering each token's attention score individually, we compare with (1) mean aggregation, which averages attention scores across all tokens into a single value, and (2) cumulative aggregation, which sums all tokens' attention scores together. 
Results show that token-wise scoring significantly outperforms these global aggregation methods. 
This performance gap can be attributed to the varying sentence lengths in prompts, which cause mean and cumulative scores to fluctuate considerably across different inputs, making it difficult to establish consistent thresholds.

\begin{table}[t!]
\centering
\resizebox{1\linewidth}{!}{
\begin{tabular}{c|ccc}
\hline
Scores & LAION-aesthetic & PickScore     & ImageReward   \\ \hline
SD3    & 6.362±0.128   & 21.812±0.487 & 0.959±0.204 \\ 
R2F    & \textbf{6.395±0.166}  & 22.157±0.301 & 0.997±0.297 \\
Ours   & 6.332±0.168  & \textbf{22.188±0.268} &  \textbf{1.039±0.260} \\ \hline
\end{tabular}
}
\caption{Aesthetic scores of generated images evaluated on RareBench. Our method shows superior PickScore and ImageReward, both capturing human-preference metrics, while maintaining comparable LAION-aesthetic scores.}
\label{tab:aesthetic}
\end{table}

\noindent \textbf{Quantitative Image Quality Analysis}
For quantitative image quality evaluation, we employ three complementary metrics: LAION-Aesthetic~\cite{schuhmann2022laion}, PickScore~\cite{kirstain2023pick}, and ImageReward~\cite{xu2023imagereward}.
We assess images generated by R2F~\cite{park2024rare}, SD3.0, and our method using all three evaluators, evaluating all methods across the full RareBench benchmark.
As shown in Tab.~\ref{tab:aesthetic}, our approach achieves the highest scores on both ImageReward and PickScore—indicating improved semantic consistency and alignment with human preferences—while maintaining a LAION-Aesthetic score comparable to R2F.

\section{Conclusion}

We present \emph{ADAPT}, a training-free framework that addresses key limitations in rare compositional concept generation.
ADAPT mitigates prompt scheduling variance and suboptimal guidance through three complementary components:
(1) Adaptive Prompt Scheduling (APS) that removes GPT-4o dependency on prompt scheduling via attention-based scheduling,
(2) Pooled Embedding Manipulation (PEM) that offers rare-specific structural guidance through orthogonal projection, and
(3) Latent Space Manipulation (LSM) that enables fine-grained attribute control.
Extensive experiments on \textit{RareBench} demonstrate that ADAPT consistently outperforms existing methods across all categories, effectively handling complex multi-object compositions while preserving visual fidelity and semantic alignment.
Overall, ADAPT establishes a deterministic and semantically grounded paradigm for rare concept generation in text-to-image synthesis.

\section*{Acknowledgments}
This was partly supported by the Institute of Information \& Communications Technology Planning \& Evaluation (IITP) grant funded by the \grantsponsor{Korean government (MSIT)} (No. \grantnumber{1}{RS-2020-II201373}, Artificial Intelligence Graduate School Program(Hanyang University)) and the Institute of Information \& Communications Technology Planning \& Evaluation (IITP) grant funded by the \grantsponsor{Korea government (MSIT)} (No. \grantnumber{2}{RS-2025-02219062}, Self-training framework for VLM-based defect detection and explanation model in manufacturing process).

{
    \small
    \bibliographystyle{ieeenat_fullname}
    \bibliography{main}

@String(TOG= {ACM Trans. Graph.})

@String(AAAI = {AAAI})

@String(TOG   = {ACM TOG})

@article{park2024rare,
  title={Rare-to-Frequent: Unlocking Compositional Generation Power of Diffusion Models on Rare Concepts with LLM Guidance},
  author={Park, Dongmin and Kim, Sebin and Moon, Taehong and Kim, Minkyu and Lee, Kangwook and Cho, Jaewoong},
  journal={arXiv preprint arXiv:2410.22376},
  year={2024}
}

@article{hurst2024gpt,
  title={Gpt-4o system card},
  author={Hurst, Aaron and Lerer, Adam and Goucher, Adam P and Perelman, Adam and Ramesh, Aditya and Clark, Aidan and Ostrow, AJ and Welihinda, Akila and Hayes, Alan and Radford, Alec and others},
  journal={arXiv preprint arXiv:2410.21276},
  year={2024}
}

@inproceedings{esser2024scaling,
  title={Scaling rectified flow transformers for high-resolution image synthesis},
  author={Esser, Patrick and Kulal, Sumith and Blattmann, Andreas and Entezari, Rahim and M{\"u}ller, Jonas and Saini, Harry and Levi, Yam and Lorenz, Dominik and Sauer, Axel and Boesel, Frederic and others},
  booktitle={Forty-first international conference on machine learning},
  year={2024}
}

@inproceedings{rombach2022high,
  title={High-resolution image synthesis with latent diffusion models},
  author={Rombach, Robin and Blattmann, Andreas and Lorenz, Dominik and Esser, Patrick and Ommer, Bj{\"o}rn},
  booktitle={Proceedings of the IEEE/CVF conference on computer vision and pattern recognition},
  pages={10684--10695},
  year={2022}
}

@article{podell2023sdxl,
  title={Sdxl: Improving latent diffusion models for high-resolution image synthesis},
  author={Podell, Dustin and English, Zion and Lacey, Kyle and Blattmann, Andreas and Dockhorn, Tim and M{\"u}ller, Jonas and Penna, Joe and Rombach, Robin},
  journal={arXiv preprint arXiv:2307.01952},
  year={2023}
}

@article{chen2023pixart,
  title={Pixart-$\alpha $: Fast training of diffusion transformer for photorealistic text-to-image synthesis},
  author={Chen, Junsong and Yu, Jincheng and Ge, Chongjian and Yao, Lewei and Xie, Enze and Wu, Yue and Wang, Zhongdao and Kwok, James and Luo, Ping and Lu, Huchuan and others},
  journal={arXiv preprint arXiv:2310.00426},
  year={2023}
}

@article{rassin2023linguistic,
  title={Linguistic binding in diffusion models: Enhancing attribute correspondence through attention map alignment},
  author={Rassin, Royi and Hirsch, Eran and Glickman, Daniel and Ravfogel, Shauli and Goldberg, Yoav and Chechik, Gal},
  journal={Advances in Neural Information Processing Systems},
  volume={36},
  pages={3536--3559},
  year={2023}
}

@article{lian2023llm,
  title={Llm-grounded diffusion: Enhancing prompt understanding of text-to-image diffusion models with large language models},
  author={Lian, Long and Li, Boyi and Yala, Adam and Darrell, Trevor},
  journal={arXiv preprint arXiv:2305.13655},
  year={2023}
}

@article{hu2024ella,
  title={Ella: Equip diffusion models with llm for enhanced semantic alignment},
  author={Hu, Xiwei and Wang, Rui and Fang, Yixiao and Fu, Bin and Cheng, Pei and Yu, Gang},
  journal={arXiv preprint arXiv:2403.05135},
  year={2024}
}

@inproceedings{samuel2024generating,
  title={Generating images of rare concepts using pre-trained diffusion models},
  author={Samuel, Dvir and Ben-Ari, Rami and Raviv, Simon and Darshan, Nir and Chechik, Gal},
  booktitle={Proceedings of the AAAI Conference on Artificial Intelligence},
  volume={38},
  number={5},
  pages={4695--4703},
  year={2024}
}

@inproceedings{radford2021learning,
  title={Learning transferable visual models from natural language supervision},
  author={Radford, Alec and Kim, Jong Wook and Hallacy, Chris and Ramesh, Aditya and Goh, Gabriel and Agarwal, Sandhini and Sastry, Girish and Askell, Amanda and Mishkin, Pamela and Clark, Jack and others},
  booktitle={International conference on machine learning},
  pages={8748--8763},
  year={2021},
  organization={PmLR}
}

@article{raffel2020exploring,
  title={Exploring the limits of transfer learning with a unified text-to-text transformer},
  author={Raffel, Colin and Shazeer, Noam and Roberts, Adam and Lee, Katherine and Narang, Sharan and Matena, Michael and Zhou, Yanqi and Li, Wei and Liu, Peter J},
  journal={Journal of machine learning research},
  volume={21},
  number={140},
  pages={1--67},
  year={2020}
}

@article{dalva2024fluxspace,
  title={FluxSpace: Disentangled Semantic Editing in Rectified Flow Transformers},
  author={Dalva, Yusuf and Venkatesh, Kavana and Yanardag, Pinar},
  journal={arXiv preprint arXiv:2412.09611},
  year={2024}
}

@inproceedings{wang2025precise,
  title={Precise, fast, and low-cost concept erasure in value space: Orthogonal complement matters},
  author={Wang, Yuan and Li, Ouxiang and Mu, Tingting and Hao, Yanbin and Liu, Kuien and Wang, Xiang and He, Xiangnan},
  booktitle={Proceedings of the Computer Vision and Pattern Recognition Conference},
  pages={28759--28768},
  year={2025}
}

@article{chefer2023attend,
  title={Attend-and-excite: Attention-based semantic guidance for text-to-image diffusion models},
  author={Chefer, Hila and Alaluf, Yuval and Vinker, Yael and Wolf, Lior and Cohen-Or, Daniel},
  journal={ACM transactions on Graphics (TOG)},
  volume={42},
  number={4},
  pages={1--10},
  year={2023},
  publisher={ACM New York, NY, USA}
}

@inproceedings{li2023gligen,
  title={Gligen: Open-set grounded text-to-image generation},
  author={Li, Yuheng and Liu, Haotian and Wu, Qingyang and Mu, Fangzhou and Yang, Jianwei and Gao, Jianfeng and Li, Chunyuan and Lee, Yong Jae},
  booktitle={Proceedings of the IEEE/CVF conference on computer vision and pattern recognition},
  pages={22511--22521},
  year={2023}
}

@inproceedings{zhang2023adding,
  title={Adding conditional control to text-to-image diffusion models},
  author={Zhang, Lvmin and Rao, Anyi and Agrawala, Maneesh},
  booktitle={Proceedings of the IEEE/CVF international conference on computer vision},
  pages={3836--3847},
  year={2023}
}

@article{feng2023layoutgpt,
  title={Layoutgpt: Compositional visual planning and generation with large language models},
  author={Feng, Weixi and Zhu, Wanrong and Fu, Tsu-jui and Jampani, Varun and Akula, Arjun and He, Xuehai and Basu, Sugato and Wang, Xin Eric and Wang, William Yang},
  journal={Advances in Neural Information Processing Systems},
  volume={36},
  pages={18225--18250},
  year={2023}
}

@inproceedings{xie2023boxdiff,
  title={Boxdiff: Text-to-image synthesis with training-free box-constrained diffusion},
  author={Xie, Jinheng and Li, Yuexiang and Huang, Yawen and Liu, Haozhe and Zhang, Wentian and Zheng, Yefeng and Shou, Mike Zheng},
  booktitle={Proceedings of the IEEE/CVF International Conference on Computer Vision},
  pages={7452--7461},
  year={2023}
}

@inproceedings{cha2025verbdiff,
  title={VerbDiff: Text-Only Diffusion Models with Enhanced Interaction Awareness},
  author={Cha, SeungJu and Lee, Kwanyoung and Kim, Ye-Chan and Oh, Hyunwoo and Kim, Dong-Jin},
  booktitle={Proceedings of the Computer Vision and Pattern Recognition Conference},
  pages={8041--8050},
  year={2025}
}

@article{kirstain2023pick,
  title={Pick-a-pic: An open dataset of user preferences for text-to-image generation},
  author={Kirstain, Yuval and Polyak, Adam and Singer, Uriel and Matiana, Shahbuland and Penna, Joe and Levy, Omer},
  journal={Advances in neural information processing systems},
  volume={36},
  pages={36652--36663},
  year={2023}
}

@inproceedings{yang2024mastering,
  title={Mastering text-to-image diffusion: Recaptioning, planning, and generating with multimodal llms},
  author={Yang, Ling and Yu, Zhaochen and Meng, Chenlin and Xu, Minkai and Ermon, Stefano and Cui, Bin},
  booktitle={Forty-first International Conference on Machine Learning},
  year={2024}
}

@article{yi2024towards,
  title={Towards understanding the working mechanism of text-to-image diffusion model},
  author={Yi, Mingyang and Li, Aoxue and Xin, Yi and Li, Zhenguo},
  journal={Advances in Neural Information Processing Systems},
  volume={37},
  pages={55342--55369},
  year={2024}
}

@article{blackforestlabs2024flux,
  title={FLUX: A powerful tool for text generation},
  author={BlackForestLabs},
  journal={https://blackforestlabs.ai/},
  year={2024}
}

@inproceedings{
Oh2025catch,
title={CatchPhrase: EXPrompt-Guided Encoder Adaptation for Audio-to-Image Generation},
author={Oh, Hyunwoo and Cha, SeungJu and Lee, Kwanyoung and Kim, Si-Woo and Kim, Dong-Jin},
booktitle={Proceedings of the 33rd ACM International Conference on Multimedia},
year={2025},
}

@article{schuhmann2022laion,
  title={Laion-5b: An open large-scale dataset for training next generation image-text models},
  author={Schuhmann, Christoph and Beaumont, Romain and Vencu, Richard and Gordon, Cade and Wightman, Ross and Cherti, Mehdi and Coombes, Theo and Katta, Aarush and Mullis, Clayton and Wortsman, Mitchell and others},
  journal={Advances in neural information processing systems},
  volume={35},
  pages={25278--25294},
  year={2022}
}

@article{xu2023imagereward,
  title={Imagereward: Learning and evaluating human preferences for text-to-image generation},
  author={Xu, Jiazheng and Liu, Xiao and Wu, Yuchen and Tong, Yuxuan and Li, Qinkai and Ding, Ming and Tang, Jie and Dong, Yuxiao},
  journal={Advances in Neural Information Processing Systems},
  volume={36},
  pages={15903--15935},
  year={2023}
}

@article{grattafiori2024llama,
  title={The llama 3 herd of models},
  author={Grattafiori, Aaron and Dubey, Abhimanyu and Jauhri, Abhinav and Pandey, Abhinav and Kadian, Abhishek and Al-Dahle, Ahmad and Letman, Aiesha and Mathur, Akhil and Schelten, Alan and Vaughan, Alex and others},
  journal={arXiv preprint arXiv:2407.21783},
  year={2024}
}
}

\clearpage
\setcounter{page}{1}
\maketitlesupplementary

\begin{figure}[t!]
    \centering
    \includegraphics[width=0.8\linewidth]{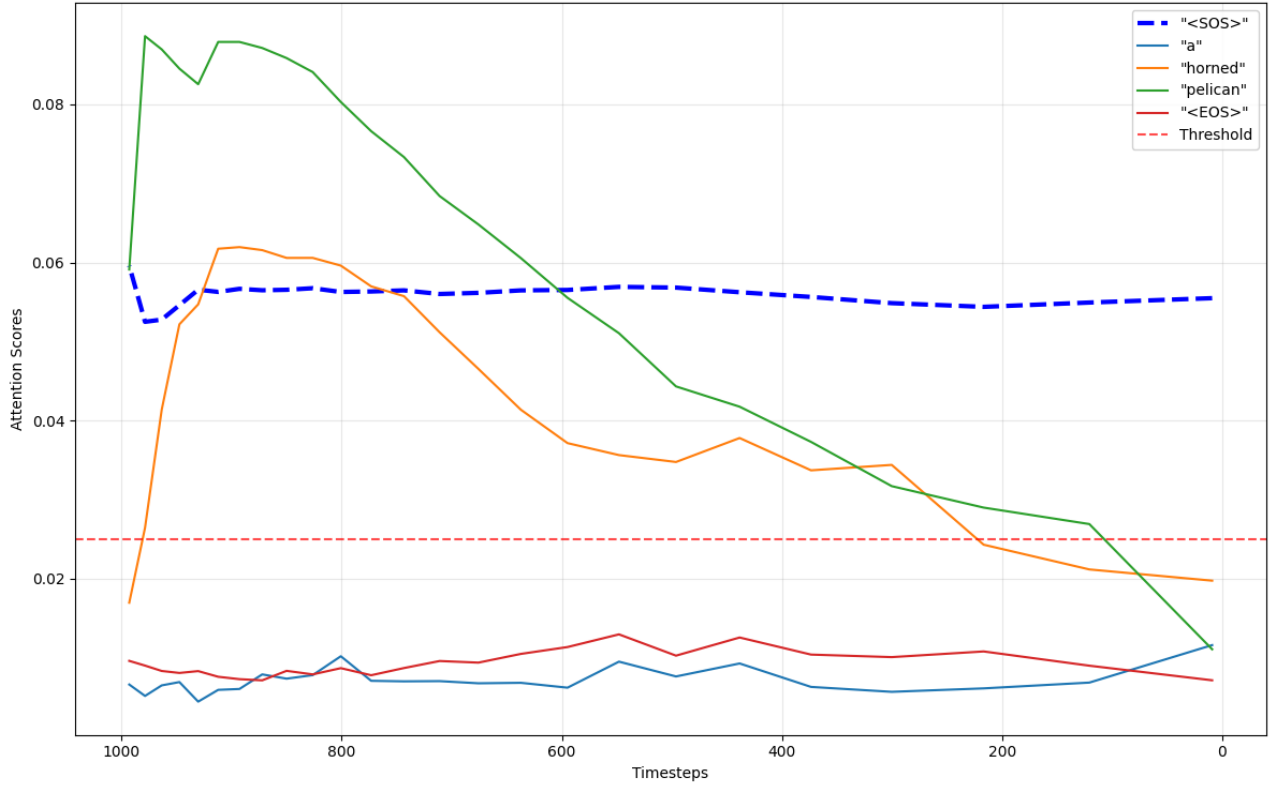}
    \caption{Attention response score $\mathcal{S}_{\text{Attn}}$ of target prompt ``A horned pelican''.}
    \label{fig:sos_token}
\end{figure}

\begin{figure*}[t!]
    \centering
    \includegraphics[width=0.8\linewidth]{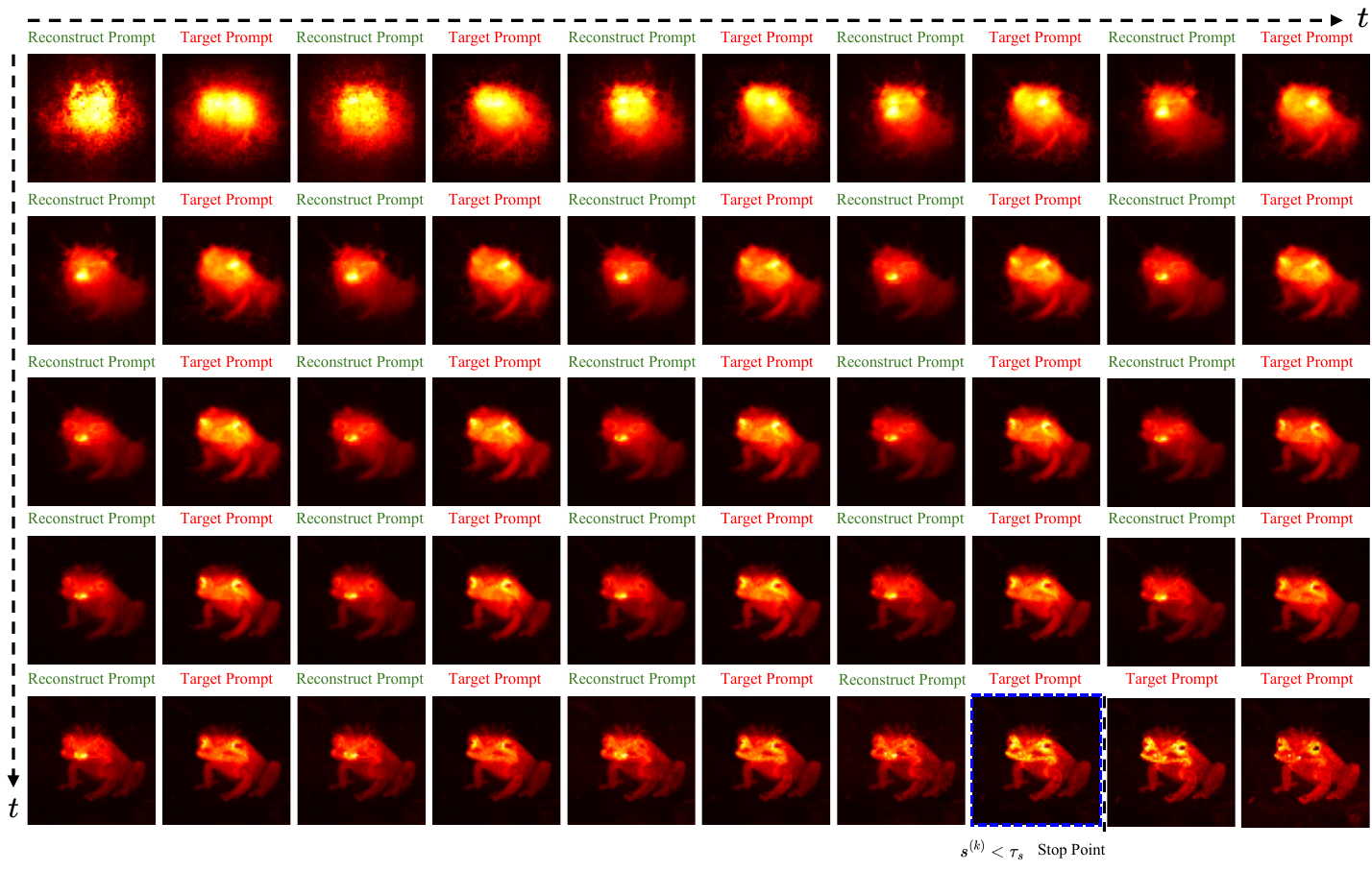}
    \caption{Attention map visualization for each step. The images are ordered sequentially from left to right across each row, starting from the top. }
    \label{fig:attention_map_ablation}
\end{figure*}

\section{Algorithm Pseudocode for Adaptive Prompt Scheduling}
\label{sec:Algorithm Pseudocode for Adaptive Prompt Scheduling}
Alg.~\ref{alg:aps} details the complete adaptive prompt scheduling procedure for ADAPT. 
The algorithm operates in two main stages: attention map scoring and prompt scheduling.
In the attention map scoring stage, for timestep $(T-t) \% 2 = 0$, we extract the maximum attention score $\mathcal{S}_{\text{Attn}}(y_{\text{tar}})$ over all tokens in the target prompt $y_{\text{tar}}$, excluding the $\langle \text{SOS} \rangle$ token.
This quantifies the attention intensity for each rare concept at the current denoising step.
If the number of transition counter $P_{\text{trans}}$ is less than the total number of rare concepts $m$, we compute the top-$k$ attention scores $s^{(k)}$ using the TopK operation. 
If the $k$-th highest attention score falls below the threshold $\tau_s$, this indicates that the semantic meaning of the token is sufficiently saturated, triggering a prompt transition. 

In the prompt scheduling stage, upon detecting low attention, we first increment the transition counter $P_\text{trans}$ to track the scheduling progress.
We then reconstruct the progressive prompt $y_{\text{prog}}$ by replacing the $P_\text{trans}$-th frequent concept ${y_f^{P_{\text{trans}}}}$ with corresponding rare concepts ${y_r^{P_{\text{trans}}}}$. 

If no transition occurs, $y_\text{prog}$ defaults to $y_\text{tar}$. 
When $(T-t) \% 2 \neq 0$, the progressive prompt $y_{\text{prog}}$ is used without scoring.
This alternation ensures gradual rare concept injection while maintaining global coherence.

\section{Why except the $<SOS>$ token?}
\label{sec:why_except_the_sos_token}

Fig.~\ref{fig:sos_token} illustrates attention score evolution for "A horned pelican."
While semantic tokens exhibit saturation patterns converging to stable values, the $<$SOS$>$ token (blue dashed line) maintains high scores throughout all timesteps.
This occurs because CLIP's causal attention prevents $<$SOS$>$ from receiving contextual information~\cite{yi2024towards}, making it a structural artifact rather than a semantic signal.
Including $<$SOS$>$ would prevent transitions in our stopping criterion as it never saturates, so we exclude it and compute $\mathcal{S}_{\text{Attn}}(y_{\text{tar}}) = \{z_i\}_{i=1}^{n}$ only over $n$ semantic tokens.

\section{Attention Map Aggregation and Scoring}

Before computing attention scores, we aggregate attention maps from all MM-DiT transformer blocks in SD3. 
For each of the $L$ transformer blocks, the attention mechanism produces an attention tensor of shape $[B, \text{num\_heads}, H, W, S]$, where $B$ is the batch size, $\text{num\_heads}$ is the number of attention heads, $H \times W$ are spatial dimensions, and $S$ is the sequence length.

We perform a two-stage aggregation process. 
First, within each block, we compute the arithmetic mean across all attention heads $\mathbf{A}^{\text{block}}_\ell=\frac{1}{|\text{num\_heads}|\sum_{h=1}^{\text{num\_heads}}}\mathbf{A}^{\text{block}}_{\ell,h}$. 
This reduces the shape from $[B, \text{num\_heads}, H, W, S]$ to $[B, H, W, S]$ for block $\ell$.

Then we stack attention maps from all $L$ blocks and compute their arithmetic mean $\mathbf{A}^c = \frac{1}{L} \sum_{\ell=1}^{L} \mathbf{A}^{\text{block}}_\ell$.
This produces the final aggregated attention map $\mathbf{A}^c \in \mathbb{R}^{[B\times H \times W \times S]}$.

From the aggregated attention map $\mathbf{A}^c$, we extract attention scores for each token in the target prompt $y_\text{tar}$.
For the $i$-th token, we compute 
$$
z_i=\max _{h, w} \mathbf{A}_{h, w, i}^c
$$
where the $\max$ operation is applied over the spatial dimensions $(h,w)$.
This yields the attention response score set $\mathcal{S}_{\text{Attn}}(y_\text{tar}) = \{ z_i\} ^n_{i=1}$, where $n$ is the number of tokens excluding the $\langle SOS\rangle$ token.
The spatial maximum operation captures the strongest attention activation for each token across the entire spatial feature map, providing a robust measure of the model's focus intensity on that token during generation.

\section{Attention Map Visualization}

Attention-based Prompt Scheduling (APS) uses attention scores to determine optimal stop points.
As shown in Fig.~\ref{fig:attention_map_ablation}, the spatial attention map sharpens after certain steps, indicating generation completion of that specific token when $s^{(k)}<\tau_s$.
Furthermore, we observe the spatial attention map pattern difference between the target and progressive prompts.
We find that this attention pattern difference hinders the convergence of the spatial attention score for the rare concept during the inference stage.

\section{Attention Scores Visualization}

\begin{figure*}
    \centering
    \includegraphics[width=1\linewidth]{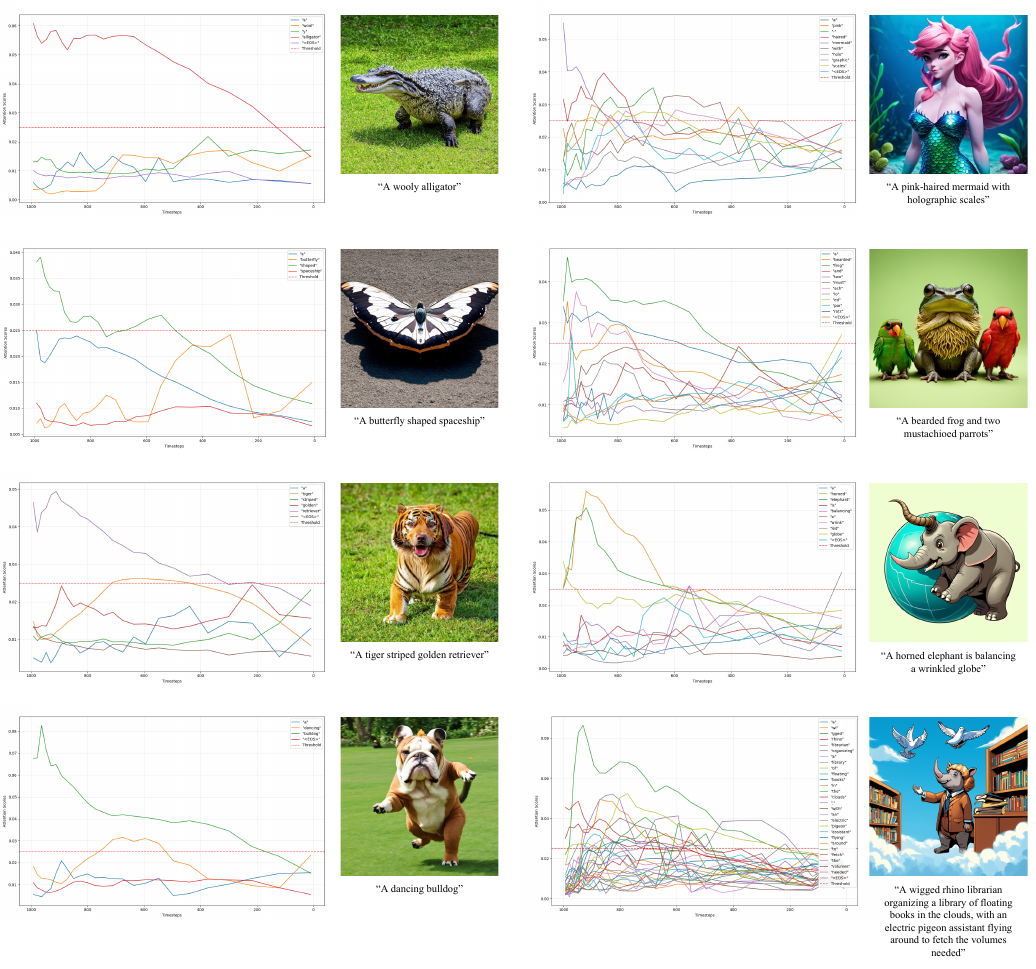}
    \caption{Visualization of $S_\text{Attn}$ on each tokens and images.}
    \label{fig:s-attn}
\end{figure*}

Figure~\ref{fig:s-attn} illustrates the temporal evolution of attention scores $S_\text{Attn}$ throughout the denoising process. For relatively simple rare concepts such as ``A wooly alligator,'' we observe that the differentiating token ``alligator'' (which distinguishes the rare prompt from its frequent counterpart ``A wooly animal'') tends to exhibit slower convergence compared to shared tokens like ``wooly.'' In such cases, the attention convergence pattern aligns with our intuition that rare semantic features require sustained focus during generation.

However, in complex multi-object or relational prompts, the convergence order becomes less predictable. Despite this variability, our APS approach using top-$k$ attention scores consistently outperforms R2F's fixed stop points across all categories. 
This suggests that even without a strict correspondence between differentiating tokens and convergence order, attention scores provide a more semantically-aware scheduling signal than predetermined timestep ratios. By adaptively determining stop points based on observed attention dynamics rather than relying on fixed heuristics, APS better aligns prompt transitions with the actual generation process.

\begin{figure*}[t]
    \centering
    \includegraphics[width=0.8\linewidth]{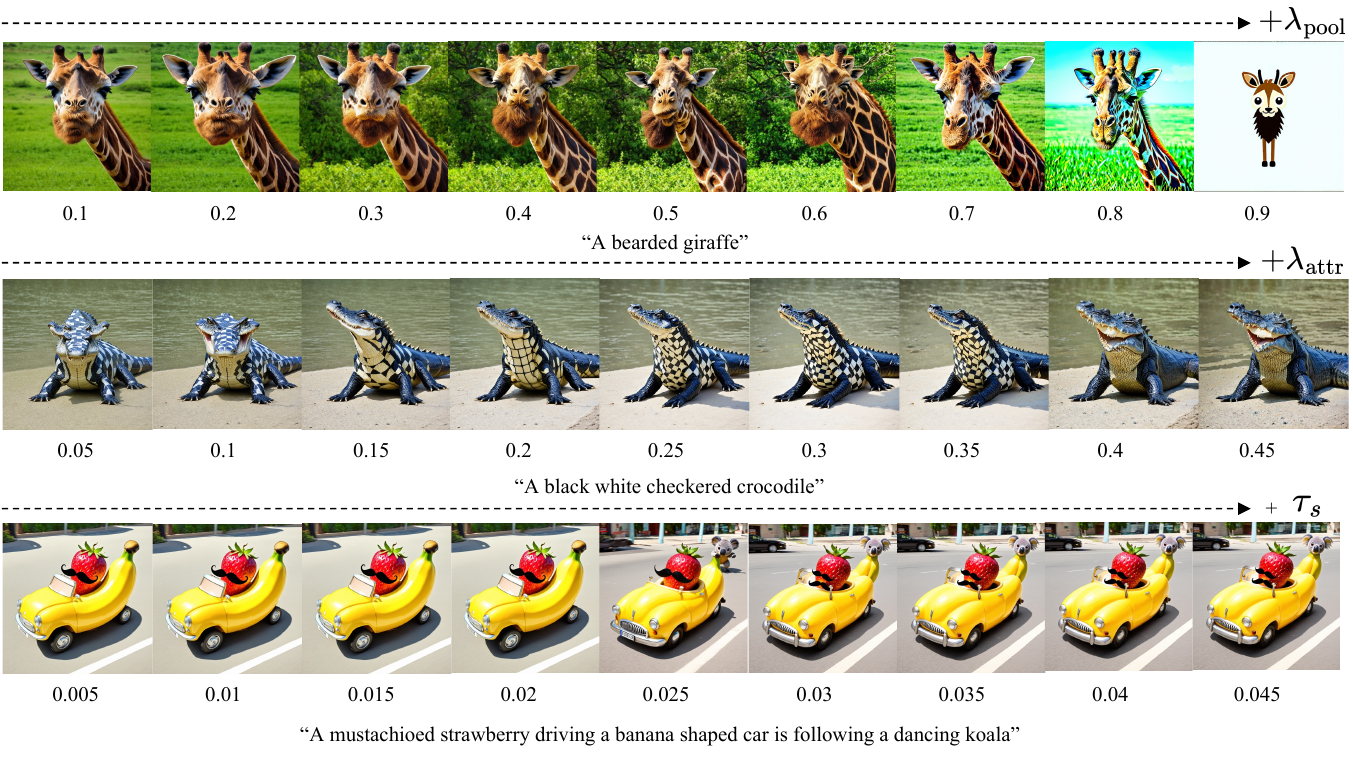}
    \caption{Further visualization on ADAPT hyperparameters for rare image generation. We vary the pooled embedding scale $\lambda_{\text{pool}}$, the latent manipulation scale $\lambda_{\text{attr}}$, and the attention threshold $\tau_s$ to analyze their effects on generation quality.}
    \label{fig:sup_hyperparams}
\end{figure*}
\begin{figure}[t!]
    \centering
    \includegraphics[width=0.8\linewidth]{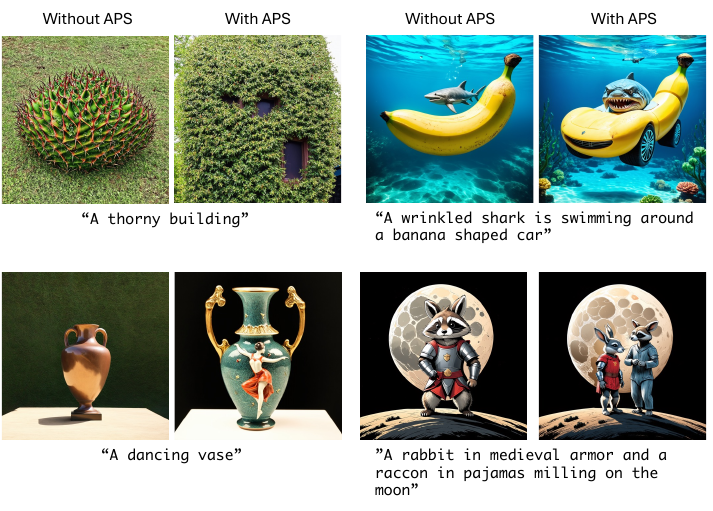}
    \caption{Ablation study with and without Attention-based Prompt Scheduling.}
    \label{fig:aps_ablation}
\end{figure}

\section{Further Visualization of Ablation Studies}
\label{sec:Further Visualization of Ablation Studies}
Fig.~\ref{fig:sup_hyperparams} shows an ablation study on ADAPT hyperparameters for rare image generation, analyzing the effects of pooled embedding scale $\lambda_\text{pool}$, latent manipulation scale $\lambda_\text{attr}$, and attention threshold $\tau_s$ on generation quality. 

Pooled embedding scale $\lambda_\text{pool}$ modulates the rare-specific disentangled semantic direction in the pooled text embedding for ``A bearded giraffe''. 
The results show that $\lambda_\text{pool}=0.3$ produces optimal results with clear beard generation while maintaining giraffe fidelity. 
When $\lambda_\text{pool} < 0.3$, the beard attribute is weakly expressed. 
However, when $\lambda_\text{pool} \geq 0.4$, image fidelity begins to degrade, and at $\lambda_\text{pool} \geq 0.7$, the rare-specific ``bearded'' attribute disappears entirely while the image becomes severely distorted, eventually reducing to a simple icon-like representation. 

Latent manipulation scale $\lambda_\text{attr}$ controls the attribute-specific disentangled semantic direction in the latent space for ``A black white checkered crocodile.'' 
The optimal result occurs at $\lambda_\text{attr}=0.15$, where the checkered pattern is clearly visible while preserving crocodile fidelity. 
When $\lambda_\text{attr} < 0.15$, the checkered pattern is insufficiently expressed.
Conversely, when $\lambda_\text{attr} > 0.15$, the crocodile structure gradually deteriorates, and at $\lambda_\text{attr} \geq 0.4$, the ``black white checkered'' attribute completely vanishes. 

Attention threshold $\tau_s$ determines the optimal stopping points $S^i$ by measuring maximum spatial attention scores for ``A mustachioed strawberry driving a banana shaped car is following a dancing koala.'' 
The results demonstrate that $\tau_s=0.025$ achieves optimal generation with all elements properly generated. 
When $\tau_s < 0.025$, some objects (particularly the koala) fail to appear or are incompletely generated. 
When $\tau_s > 0.025$, object generation becomes truncated, with only partial elements (e.g., koala head only) being produced.

\section{Impacts on Attention-based Prompt Scheduling}

Fig.~\ref{fig:aps_ablation} shows the qualitative analysis of the Adaptive Prompt Scheduling (APS), comparing results with and without APS.
Without APS, ``A thorny building,'' ``A dancing vase,'' ``A wrinkled shark is swimming around a banana shaped car,'' and ``A rabbit in medieval armor and a raccoon in pajamas milling on the moon,'' miss critical elements ``building,'' ``dancing,'' ``banana shaped car,'' and ``A rabbit'' in the generated results.
However, with APS, all prompt components are successfully incorporated.
This demonstrates the superiority of APS that ensures comprehensive semantic coverage by adaptively scheduling attention across all prompt components.

\begin{table*}[]
\centering
\begin{tabular}{p{\textwidth}}
\hline
\textbf{$<$Instructions for Image Preference Evaluation$>$} \\
In this study, you will be shown a series of image pairs, each accompanied by a text prompt describing a rare or unusual concept. For each pair, please select the image that you believe best matches the given text prompt. \\
\\
\textbf{What to Consider} \\
(1) Semantic accuracy: Does the image contain all objects and attributes mentioned in the prompt? \\
(2) Visual coherence: Is the rare concept visually well-integrated and believable?\\
\\

Please focus on how well each image represents the described rare concept, even if such concepts don't exist in reality.  The study consists of 40 prompts, selected from each category of RareBench. The entire process should only take about 20 minutes. Your input is extremely valuable and will help us understand how different image generation methods handle rare and unusual concepts. \\
\hline
\caption{User Study Instructions.}
\label{tab:user study instructions}
\end{tabular}
\end{table*}

\begin{figure}[t!]
    \centering
    \includegraphics[width=0.8\linewidth]{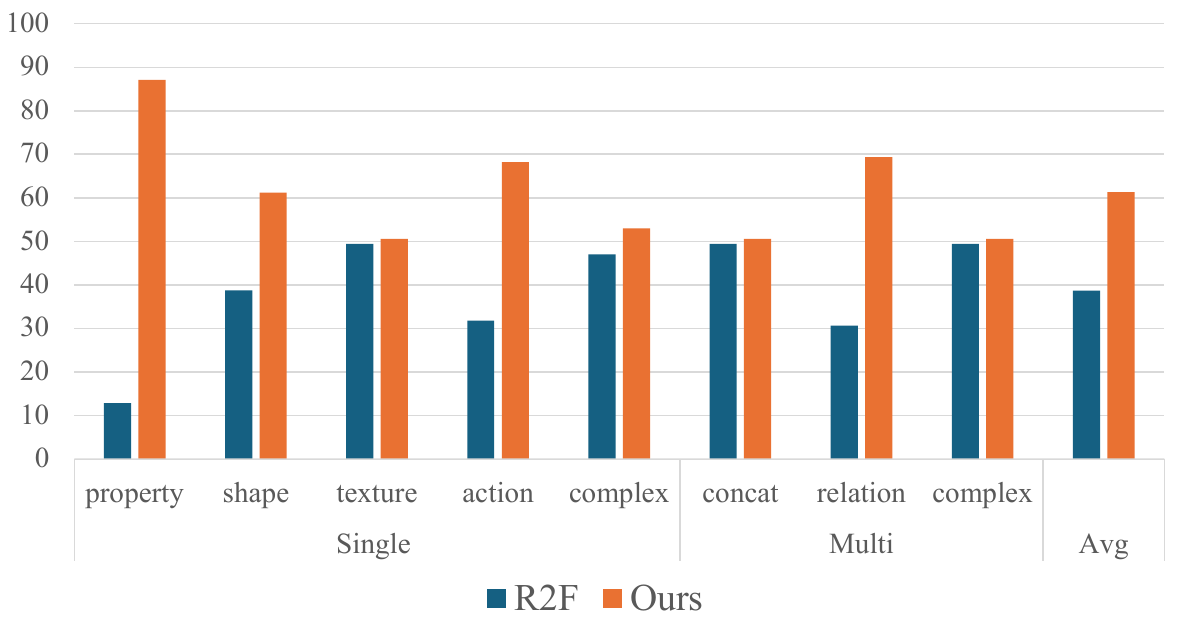}
    \caption{User study results for each category in the RareBench benchmark. Participants consistently preferred images generated by ADAPT over R2F, indicating stronger text-to-image alignment and rare concept fidelity.}
    \label{fig:user-study}
\end{figure}

\begin{figure}[t!]
    \centering
    \includegraphics[width=0.8\linewidth]{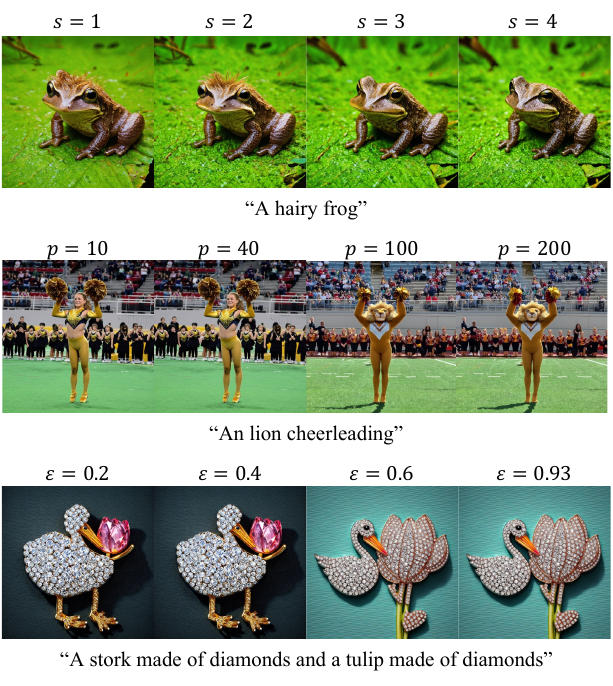}
    \caption{Qualitative analysis of adaptive weighting in PEM with varying parameters $(s, p, \epsilon)$ for scaling range, sharpness, and similarity threshold, respectively.}
    \label{fig:adaptive}
\end{figure}

\section{User Study}
\label{sec:user-study}
Fig.~\ref{fig:user-study} summarizes our user study evaluating human preferences for rare concept generation. 
We recruited 20 anonymous participants, each of whom compared 40 side-by-side image pairs generated by our method and R2F. 
The pairs were randomly drawn from RareBench, with 5 prompts sampled from each category to ensure balanced coverage. 
For every pair, participants selected the image that best matched the prompt based on two criteria—semantic accuracy and visual coherence—with image positions randomized to avoid bias. 
Overall, our method received $22.7\%$ more preferences than R2F, demonstrating consistently stronger text-to-image alignment across categories. 
Tab.~\ref{tab:user study instructions} provides the full evaluation instructions.

\section{Qualitative Results on Adaptive Weighting}

Fig.~\ref{fig:adaptive} shows the adaptive weighting hyperparameters used in the PEM.
We follow the hyperparameter settings from~\cite{wang2025precise}.
When $s \geq 3$, the ``hairy'' feature in ``A hairy frog'' disappears.
And when the $p \geq 100$, ``lion'' attribute appears on the ``An lion cheerleading.''
For the threshold $\epsilon$, values above $0.6$ produce images with correct features of ``a tulip made of diamonds.''

\begin{figure}[t!]
    \centering
    \includegraphics[width=0.9\linewidth]{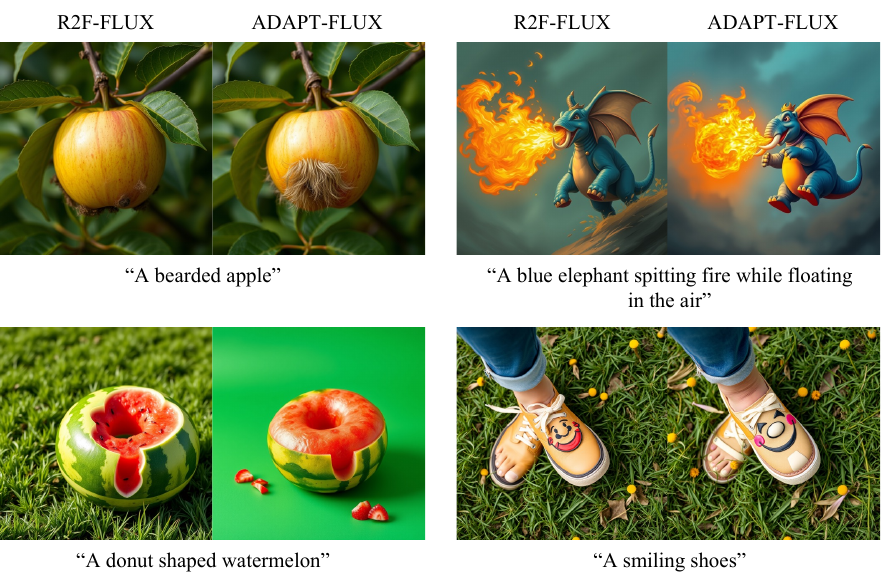}
    \caption{4-step inference qualitative results of ADAPT combined with FLUX-schnell.}
    \label{fig:flux}
\end{figure}

\begin{table}[t!]
\centering
\resizebox{\linewidth}{!}{
\begin{tabular}{l|ccccc}
\hline
RareBench    & Property & Shape & Texture & Action & Complex \\ \hline
FLUX         & 72.5     & 68.1  & 49.3    & 61.2   & 73.7    \\
R2F (FLUX)   & 78.7     & 75.0  & 56.8    & 67.5   & 68.7    \\ \hline
ADAPT (FLUX) & \textbf{81.9}     & \textbf{80.6}  & \textbf{61.3}    &\textbf{ 70.0}   & \textbf{77.5 }    \\ \hline
\end{tabular}
}
\caption{4-step inference quantitative results of ADAPT combined with FLUX-schnell.}
\label{tab:flux}
\end{table}

\section{ADAPT with FLUX-schnell}

Here, we further propose an accelerated version of ADAPT integrated with FLUX-schnell, which requires only 4 or fewer steps to generate each image. 
We maintain only the core ideas of PEM and LSM, while following R2F's prompt scheduling for FLUX-schnell integration to handle short inference steps. 
Tab.~\ref{tab:flux} shows the 4-step inference results of ADAPT combined with FLUX-schnell on RareBench.
While R2F applies the Composable method to pooled text prompts, we employ PEM for pooled text embeddings. 
To inject attribute-specific guidance, we utilize LSM within the FLUX latent space. 
Fig.~\ref{fig:flux} shows the generated images of ADAPT combined with FLUX-schnell.
Overall, ADAPT enhances text-to-image alignment while maintaining image quality, demonstrating that disentangled guidance is crucial for rare concept generation.

\section{Further Visualization}
\label{sec:Further Visualization}

\begin{figure*}
    \centering
    \includegraphics[width=1\linewidth]{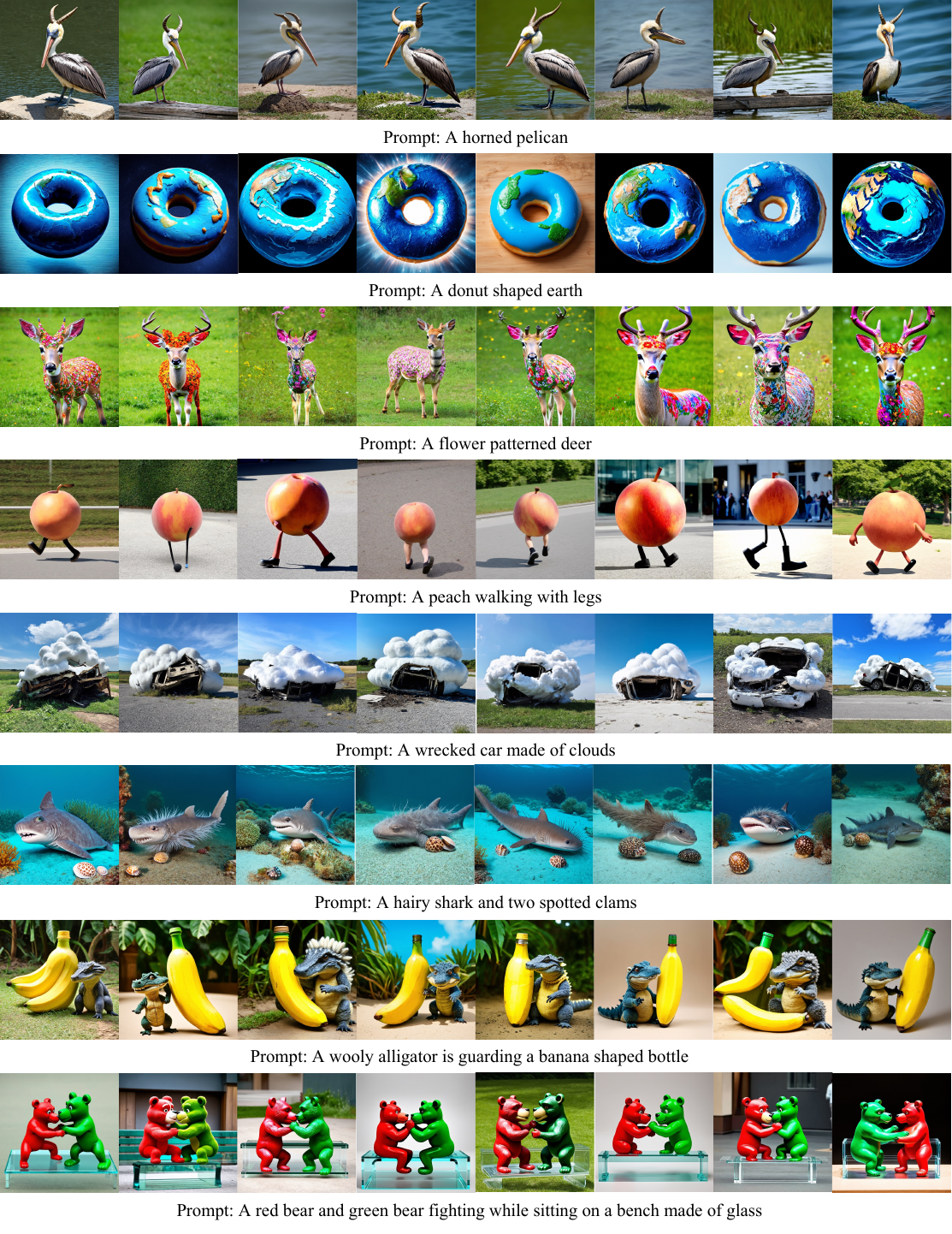}
    \caption{Random seeds visualization results of ADAPT on RareBench. Images are generated from eight random prompts with eight random seeds.}
    \label{fig:further visualization}
\end{figure*}

Fig.~\ref{fig:further visualization} shows generated images on various seeds of ADAPT on RareBench.
We randomly select eight prompts from RareBench's various categories and generate images using eight random seeds.
Overall, most of the generated images are well-aligned with the input prompt, without compromising the fidelity and quality.

\section{LLM Instruction for ADAPT}
\label{sec:LLM_Instruction_for_ADAPT}

Tab.~\ref{tab:llm_instruction} presents the complete LLM instruction and in-context examples for ADAPT. To support LSM requirements, we incorporate \textbf{Context} fields corresponding to $\{y_\text{attr}^i\}_{i=1}^m$ in the example outputs. 

The LLM decomposes input prompts into rare-to-frequent concept mappings with extracted contexts $\{y_r^i, y_f^i,y_\text{attr}^i\}_{i=1}^m$, enabling automatic rare concept identification, frequent concept generation, and attribute extraction in a one-shot LLM inference.

\begin{figure}[t!]
    \centering
    \includegraphics[width=1\linewidth]{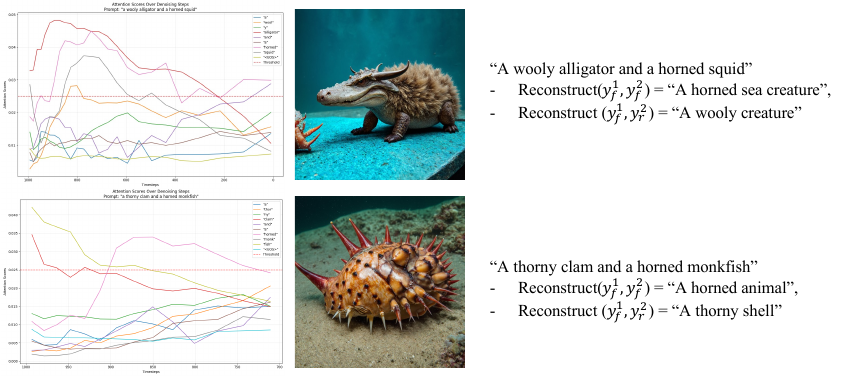}
    \caption{Illustration of failure cases. For the fair comparison, we use the same reconstructed prompts as R2F.}
    \label{fig:failure cases}
\end{figure}

\section{Frequent Prompt Selection Strategies.} 
To examine the sensitivity of ADAPT to the quality of frequent prompts, we generate three frequent prompt selection strategies on RareBench (Tab.~\ref{tab:anchor}).
The \textit{Human Generated} frequent prompt leverage expert-curated frequent prompts released by R2F authors; due to missing annotations for the complex category, we report the average over the remaining categories.
For the \textit{LLaMA3-8B-Instruct}~\cite{grattafiori2024llama} and \textit{GPT-4o} configurations, we rely on LLM-generated frequent prompts produced according to our LLM instructions.
Across all prompt selection strategies, ADAPT consistently outperforms R2F, demonstrating strong robustness to the choice and quality of frequent prompts.
Notably, frequent prompts generated by GPT-4o lead to stronger results than human-curated prompts, suggesting that LLM-assisted prompt construction can provide richer or more semantically suitable anchors for rare concept generation. 

\begin{table}[t!]
\centering
\resizebox{1\linewidth}{!}{
\begin{tabular}{c|ccccc|c}
\hline
Models          & Property & Shape & Texture & Action & Complex & Avg \\ \hline
SD3.0           & 49.4     & 76.3  & 53.1    & 71.9   & 65.0  & 63.1  \\ \hline
Human Generated (R2F) & 79.8    & 68.8  & 76.3    & 78.5   & -     & 75.9   \\
Human Generated (Ours) &\textbf{92.5}    & \textbf{83.8}  & \textbf{81.9 }   & \textbf{79.9}   & -     & \textbf{84.5}  \\ \hdashline
LLaMA3 (R2F)          & \textbf{81.9 }   & 77.1  & \textbf{76.3}    & \textbf{78.8}   & 67.7  & 76.4 \\
LLaMA3 (Ours)         & 78.8     & \textbf{80.6} & \textbf{76.3}   & 78.1   & \textbf{70.0}  & \textbf{76.8} \\ \hdashline
GPT-4o (R2F)             & 89.4     & 79.4  & 81.9    & 80.0   & 72.5  & 80.6  \\ 
GPT-4o (Ours)        & \textbf{96.3}     & \textbf{88.8}  & \textbf{83.8}    &\textbf{81.9}   & \textbf{79.4}  & \textbf{86.0}  \\ \hline
\end{tabular}
}
\caption{Experiments with different frequent prompt strategies.}
\label{tab:anchor}
\end{table}

\section{Limitation}

Fig.~\ref{fig:failure cases} shows that when reconstructed and target prompts differ greatly in length or semantics, the prompt schedule becomes challenging due to insufficient attention saturation.
While our framework demonstrates robust performance across a wide range of prompt pairs, a possible extension is to generate frequent prompts that are semantically aligned and length-consistent with the target prompt, ensuring more stable scheduling.

\clearpage

\begin{table*}[]
\centering

\begin{tabular}{p{\textwidth}}
\hline
\textbf{$<$System Prompt$>$} \\
You are a helper language model for a text-to-image generation program that aims to create images based on input text. The program often struggles to accurately generate images when the input text contains rare concepts that are not commonly found in reality. To address this, when a rare concept is identified in the input text, you should replace it with relevant yet more frequent concepts. \\
\\
\textbf{$<$User Prompt$>$} \\
Extract rare concepts from the input text and replace them with relevant yet more frequent ones. Perform the following process step by step: \\
a. Identify and extract any rare concepts from the provided input text. If the text contains one or more rare concepts, extract them all. If there are no rare concepts present, do not extract any concepts. The extracted rare concepts should not overlap. \\
b. Given the rare concepts, replace each extracted rare concept with a more frequent concept. Specifically, split each rare concept into the main noun subject and the context, and replace the main noun subject with a more frequent noun subject that is likely to appear in the context of the original rare concept. \\
c. Generate a text sequence that starts from the text with replaced frequent concepts and ends with the text with the original rare concepts. \\
The output should follow the format of the examples below: \\
\\
\textbf{$<$In-context Examples$>$} \\
\textbf{Input:} A peach made of glass \\
\textbf{Output:} \\
\textbf{Num Rare Concepts:} 1 \\
a. Rare concept: A peach made of glass \\
b. A peach made of glass does not exist in reality, while the possibility of a pink sphere made of glass existing is much higher. Main noun subject: peach, Context: made of glass, Replaced frequent subject: pink sphere \\
c. A pink sphere made of glass BREAK A peach made of glass \\
\textbf{Context:} made of glass \\
\textbf{Final Prompt Sequence:} A pink sphere made of glass BREAK A peach made of glass \\
\\
\textbf{Input:} A horned frog \\
\textbf{Output:} \\
\textbf{Num Rare Concepts:} 1 \\
a. Rare concept: A horned frog \\
b. A horned frog does not exist in reality, while a horned animal does. Main noun subject: frog, Context: a horned, Replaced frequent subject: animal \\
c. A horned animal BREAK A horned frog \\
\textbf{Context:} a horned \\
\textbf{Final Prompt Sequence:} A horned animal BREAK A horned frog \\
\\
\textbf{Input:} A horned lion and a hairy frog \\
\textbf{Output:} \\
\textbf{Num Rare Concepts:} 2 \\
a. Rare concept: A horned lion \\
b. A horned lion does not exist in reality, while a horned animal does. Main noun subject: lion, Context: horned, Replaced frequent subject: animal \\
c. A horned animal BREAK A horned lion \\
\textbf{AND} \\
a. Rare concept: A hairy frog \\
b. A hairy frog does not exist in reality, while a hairy animal does. Main noun subject: frog, Context: a hairy, Replaced frequent subject: animal \\
c. A hairy animal BREAK A hairy frog \\
\textbf{Context:} horned AND a hairy \\
\textbf{Final Prompt Sequence:} A horned animal BREAK A horned lion AND A hairy animal BREAK A hairy frog \\
\hline
\caption{Full LLM instruction for ADAPT to generate rare-to-frequent concept mappings.}
\label{tab:llm_instruction}
\end{tabular}
\end{table*}

\end{document}